\pdfoutput=1

\documentclass[11pt]{article}

\usepackage{acl}

\usepackage{times}
\usepackage{latexsym}

\usepackage[T1]{fontenc}

\usepackage[utf8]{inputenc}

\usepackage{microtype}

\usepackage{amsmath}
\usepackage{multirow}
\usepackage{listings}
\usepackage{color}
\usepackage{graphicx}
\usepackage{subcaption}
\usepackage{tcolorbox}
\usepackage{makecell}
\usepackage{booktabs}
\usepackage{cleveref}
\usepackage{colortbl}
\usepackage{soul}

\definecolor{dkgreen}{rgb}{0,0.6,0}
\definecolor{gray}{rgb}{0.5,0.5,0.5}
\definecolor{mauve}{rgb}{0.58,0,0.82}

\definecolor{dgreen}{rgb}{0.412,0.741,0.271}
\definecolor{dblue}{rgb}{0.220,0.325,0.639}
\definecolor{dred}{rgb}{0.933,0.122,0.137}

\definecolor{g1}{HTML}{b3e2cd}
\definecolor{r1}{HTML}{fdcdac}
\definecolor{w1}{HTML}{cbd5e8}
\definecolor{b1}{HTML}{fff7bc}

\definecolor{lr}{HTML}{bebada}
\definecolor{fr}{HTML}{fccde5}

\newcommand{\ie}{\textit{i}.\textit{e}.\,}
\newcommand{\eg}{\textit{e}.\textit{g}.\,}

\newcommand{\centerone}[2]{\multicolumn{1}{>{\columncolor{#1}}c}{#2}}
\newcommand{\centeroneg}[1]{\centerone{g1}{#1}}
\newcommand{\centeroner}[1]{\centerone{r1}{#1}}

\newcommand{\centeroneb}[1]{\centerone{b1}{#1}}

\newcommand{\centerthree}[2]{\multicolumn{3}{>{\columncolor{#1}}c}{#2}}
\newcommand{\centerthreeg}[1]{\centerthree{g1}{#1}}
\newcommand{\centerthreer}[1]{\centerthree{r1}{#1}}

\newcommand{\centerthreeb}[1]{\centerthree{b1}{#1}}

\newtcbox{\entoure}[1][blue]{on line,
arc=1.5pt,colback=#1!8!white,colframe=#1!32!white,
before upper={\rule[-1pt]{0pt}{10pt}},boxrule=.5pt,
boxsep=0pt,left=2pt,right=2pt,top=1pt,bottom=.5pt}

\newcommand\blfootnote[1]{%
  \begingroup
  \renewcommand\thefootnote{}\footnote{#1}%
  \addtocounter{footnote}{-1}%
  \endgroup
}

\title{QRelScore: Better Evaluating Generated Questions with Deeper Understanding of Context-aware Relevance}

\author{
    Xiaoqiang Wang\textsuperscript{\rm 1$*$},
    Bang Liu\textsuperscript{\rm 2$*$$\dagger$},
    Siliang Tang\textsuperscript{\rm 1$\ddagger$} \and
    Lingfei Wu\textsuperscript{\rm 3$\ddagger$} \\
    \textsuperscript{\rm 1}Zhejiang University, \textsuperscript{\rm 2}Universit{\'e} de Montr{\'e}al \& Mila \\
     \textsuperscript{\rm 3}JD.COM Silicon Valley Research Center \\
    \{\texttt{xq.wang, siliang}\}\texttt{@zju.edu.cn} \\
    \texttt{bang.liu@umontreal.ca} \\
    \texttt{lwu@email.wm.edu} \\
}

\begin{document}
\maketitle
\begin{abstract}
Existing metrics for assessing question generation not only require costly human reference but also fail to take into account the input context of generation, rendering the lack of deep understanding of the relevance between the generated questions and input contexts.
As a result, they may wrongly penalize a legitimate and reasonable candidate question when it (\romannumeral 1) involves complicated reasoning with the context or (\romannumeral 2) can be grounded by multiple evidences in the context.
In this paper, we propose \textbf{QRelScore}, a context-aware \underline{\textbf{Rel}}evance evaluation metric for \underline{\textbf{Q}}uestion Generation.
Based on off-the-shelf language models such as BERT and GPT2, QRelScore employs both word-level hierarchical matching and sentence-level prompt-based generation to cope with the complicated reasoning and diverse generation from multiple evidences, respectively.
Compared with existing metrics, our experiments demonstrate that QRelScore is able to achieve a higher correlation with human judgments while being much more robust to adversarial samples.
\vspace{-8mm}
\end{abstract}

\blfootnote{$^*$Equal contribution.}
\blfootnote{$^\dagger$Canada CIFAR AI Chair.}
\blfootnote{$^\ddagger$Corresponding authors.}

\section{Introduction}
\label{sec:introduction}

\begin{table*}[!t]
    \setlength\tabcolsep{2pt}
    \resizebox{1.0\textwidth}{!}{
        \begin{tabular}{lcccc}
        \toprule
        \multicolumn{5}{l}{\textbf{Context.} ...\textcolor{dgreen}{in 1987}, when some students believed that the observer began to show a conservative bias, a liberal newspaper,} \\
        
        \multicolumn{5}{l}{Common Sense was published...} \\
        \hline

        \textbf{Reference.} when was Common Sense published for the first time?& \textbf{BLEU1}& \textbf{ROUGE-L}& \textbf{Q-BLEU1}& \textbf{BERTScore} \\
        \hline
        \textbf{$Q_1$ Candidate.} when was Common Sense first published?& 0.651& 0.6841& 0.800& 0.791 \\

        \textbf{$Q_2$ Unanswerable.} who was Common Sense published for the first time?& \textcolor{dred}{0.899}& \textcolor{dred}{0.900}& 0.417& \textcolor{dred}{0.998} \\

        \textbf{$Q_3$ Paraphrasing.} in what year did Common Sense begin publication?& \entoure{0.181}& \entoure{0.192}&  \entoure{0.276}& 0.671 \\

        \textbf{$Q_4$ Coreference.} in what year did the student liberal newspaper begin publication?& \entoure{0.181}& \entoure{0.192}&  \entoure{0.053}& \entoure{0.291} \\

        \textbf{$Q_5$ Other evidences.} when did the observer begin to show a conservative bias?& \entoure{0.272}& \entoure{0.288}& \entoure{0.427}& \entoure{0.265} \\

        \bottomrule
        \end{tabular}
    }
    \caption{
    Five generated questions, the context, the ground-truth answer span (colored in \textcolor{dgreen}{green}) that the question is generated for, and the human reference.
    We \entoure{box} the cases where the well-formed and meaningful candidates are scored much lower than the candidate $Q_1$.
    In contrast, the unanswerable adversarial example with a higher score than the candidate $Q_1$ is marked in \textcolor{dred}{red}.
    In other words, they reflect those previous metrics wrongly score the candidates too low or high, respectively.
    }
    \label{tab:intro-examples}
    \vspace{-5mm}
\end{table*}

Question generation (QG) systems aim to generate natural language questions that are relevant to and usually can be answered by a given piece of input text~\cite{chen2019reinforcement,liu2020asking,liu2019learning}.
As an important natural language generation task, QG can be used to improve various applications, such as question answering (QA)~\cite{fabbri-etal-2020-template, yu-etal-2020-technical, cheng-etal-2021-guiding}, conversational systems~\cite{wang-etal-2018-learning-ask,chen2019graphflow}, and information retrieval (IR)~\cite{yu-etal-2020-review}.
Meanwhile, it has long been criticized that QG models usually suffer from the semantic drift problem owing to the widely adopted likelihood-based training, \ie the models ask questions that are not relevant to and can not be supported by the context~\cite{zhang-bansal-2019-addressing,chen2020toward}.
Thus, how to accurately evaluate the relevance between generated questions and the context is attracting more and more attention.
One of the most accurate evaluation methods is human evaluation.
However, human evaluation is expensive, time-consuming, and non-reproducible.
Therefore, it is necessary to develop automatic evaluation metrics for question generation systems.

Traditional automatic metrics (\eg BLEU~\cite{papineni-etal-2002-bleu}, ROUGE~\cite{lin-2004-rouge} and METEOR~\cite{banerjee-lavie-2005-meteor}) measure the n-gram overlap between the candidate and corresponding reference question, but they often fail to robustly match paraphrases.
More recently, Q-BLEU~\cite{nema-khapra-2018-towards} and BERT-based metrics such as BERTScore~\cite{zhang2019bertscore}, MoverScore~\cite{zhao-etal-2019-moverscore} and LS\_Score~\cite{wu-etal-2020-unsupervised} were proposed to evaluate the answerability and semantic similarity of a candidate question, achieving better correlation with human judgments.

However, they compute the similarity between the system output and the reference without considering the crucial input context of generation.
Therefore, they cannot properly capture the reasoning relationship between the generated output and input context.
Furthermore, comparing with a reference question omits the incompleteness of the reference: we can ask different questions based on the same context by paying attention to different information (or evidence) in it, while the reference question only represents one possible output.
As a result, existing QG or text generation metrics struggle in evaluating the quality of candidate questions that (\romannumeral 1) involve complicated reasoning with the context, or (\romannumeral 2) are generated from the evidence in the context that differs from the reference questions.


Table~\ref{tab:intro-examples} shows an example of question generation that exemplifies some weaknesses of previous metrics.
BLEU1 and ROUGE-L cannot detect the unanswerable question ($Q_2$) and wrongly score the other well-formed candidates ($Q_3$ - $Q_5$) significantly lower than the candidate $Q_1$.
Although Q-BLEU1 successfully penalizes the unanswerable question, it fails to discern the complicated but beneficial paraphrasing candidate ($Q_3$).
BERTScore leverages contextualized embeddings and shows some degree of ability to distinguish the paraphrasing candidate, but it cannot perform linguistic reasoning related to the context (such as coreference resolution for $Q_4$) and scores the legitimate novel generation from other evidence ($Q_5$) much lower than the candidate $Q_1$.


In this paper, we present \textit{QRelScore}, an automatic reference-free evaluation metric for question generation (QG), which requires neither supervision from human ratings nor additional training on specific domains.
QRelScore addresses the weaknesses above by considering the context-aware relevance in a word- and sentence-level manner.
On the one hand, inspired by the hierarchical procedure taken by masked language models such as BERT to understand a question~\cite{van2019does}, QRelScore understands the word-level relevance by explicitly capturing the reasoning relationship between the candidate tokens and the context tokens.
On the other hand, based on the benefit of intra-sentence coherence in the autoregressive language models such as GPT2 that originates from the word-by-word nature of human language production, the sentence-level relevance is measured by the overall factual consistency between the candidate and all the possible evidences in the context.

We verify the effectiveness and efficiency of QRelScore through various experiments.
First, we demonstrate that QRelScore can improve the performance of question answering: by serving as a reward to train a QG model with reinforcement learning and then use it to augment a QA dataset (\eg the SQuAD dataset~\cite{rajpurkar-etal-2016-squad}), the performance of a QA model can be improved by fine-tuning on the augmented dataset.
Second, our QRelScore achieves a state-of-the-art correlation with human judgments on the candidates generated by the existing QG models.
Furthermore, when considering the available human reference of the dataset in our QRelScore, we present a reference-augmented version, Ref-QRelScore, which achieves an even higher correlation.
Last, extensive experiments on the robustness test also demonstrate that QRelScore has a stronger ability to discriminate against adversarial samples when compared to existing metrics.

\section{QRelScore Metric}
\label{sec:metric}

In this section, we formulate our reference-free evaluation metric QRelScore based on the off-the-shelf pre-trained language models.
Specifically, QRelScore consists of two scoring components: the local relevance matching (QRel$_{LRM}$) component and the global relevance generation (QRel$_{GRG}$) component.
The former is used to handle the candidates involving complicated reasoning with the contexts by computing word-level similarity using layer-wise embeddings and cross attention, while the latter is responsible for measuring the factual consistency between the candidate and all evidences of a given answer by comparing the difference in the confidence of generating the context with or without a prompt.

\begin{figure}[!t]
    \includegraphics[width=\columnwidth]{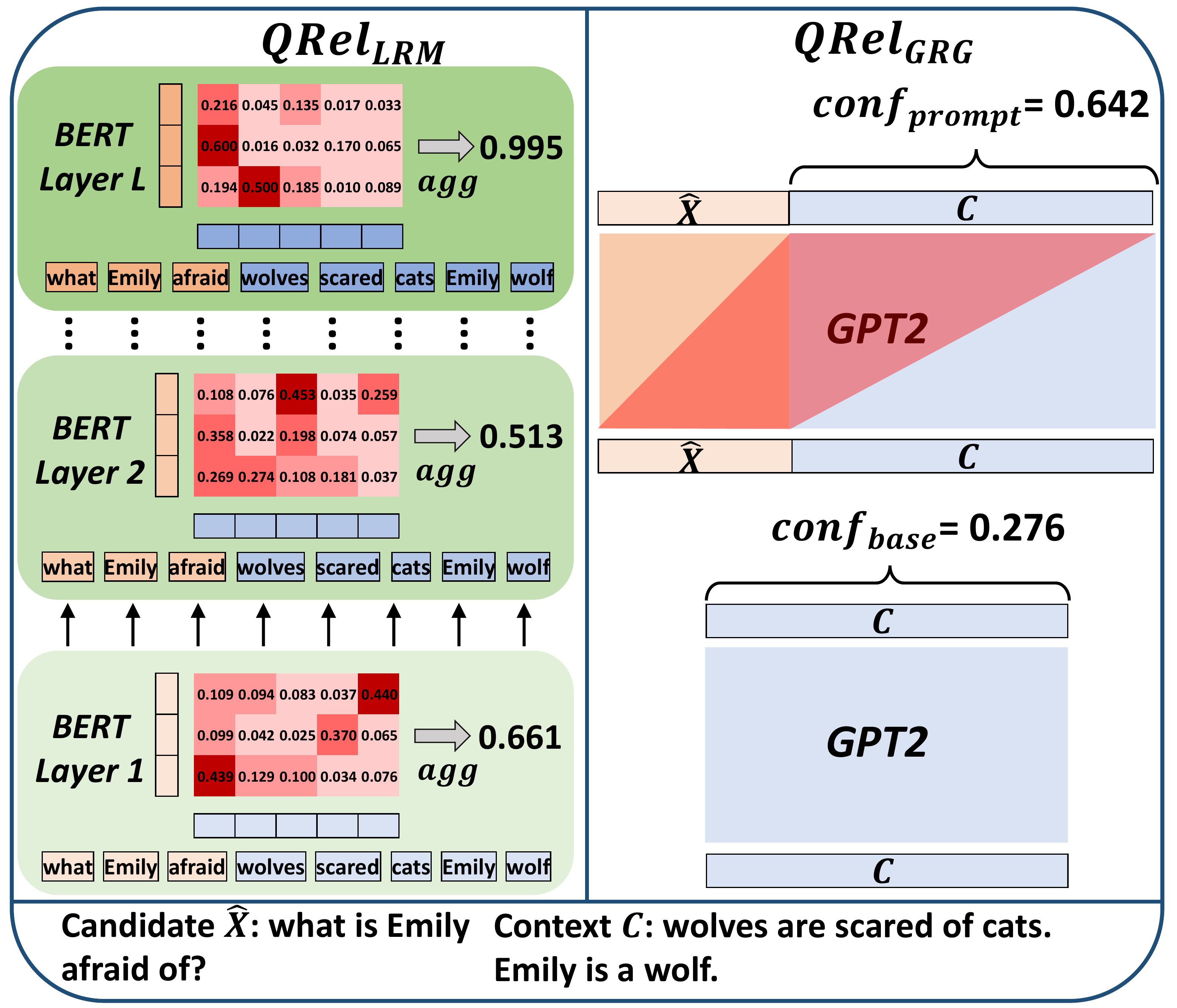}
    \caption{ Illustration of the computation of our $\text{QRel}_{LRM}$ (\emph{left} part) and $\text{QRel}_{GRG}$ (\emph{right} part).
    $\text{QRel}_{LRM}$ is based on the layer-wise cross attention by feeding the candidate and context together into BERT, while $\text{QRel}_{GRG}$ is formulated as the confidence gain obtained by employing the candidate as a prompt (\textcolor{dred}{red} region in the \emph{right} part) of GPT2.
    Several stop words are omitted in the representation of attention maps.
    }
    \label{fig:qrelscore}
    \vspace{-5mm}
\end{figure}

Figure~\ref{fig:qrelscore} illustrates the computation of QRel$_{LRM}$ and QRel$_{GRG}$.
Given a candidate question $\hat{X} = \langle \hat{x}_1, \cdots, \hat{x}_m, \cdots, \hat{x}_M \rangle $ and its context $C = \langle c_1, \cdots, c_n, \cdots, c_N \rangle$, our QRelScore is computed as the harmonic mean of QRel$_{LRM}$ and QRel$_{GRG}$:
\begin{equation}
    \text{QRelScore}(\hat{X}, C) = 2\frac{\text{QRel}_{LRM} \cdot \text{QRel}_{GRG}}{\text{QRel}_{LRM} + \text{QRel}_{GRG}}
\end{equation}

\subsection{Local Relevance Matching}
QRel$_{LRM}$ computes the word-level similarity between the candidate and context using the layer-wise contextualized embeddings and the cross attention between them.
Although the contextualized embeddings have been introduced in the evaluation of the text generation task, there are two critical differences in its utilization between our QRel$_{LRM}$ and previous works such as BERTScore~\cite{zhang2019bertscore} and Moverscore~\cite{zhao-etal-2019-moverscore}.
First, we feed the candidate and the context into the model together, whereas previous works feed them in a 2-step division, first for the candidate and then for the context.
Therefore, we can make full use of the available cross attention between the candidate and the context to weigh the importance of every token better than previous works, whose weighting is based on the hand-crafted inverse document frequency (IDF).
Because the IDF weighting only considers the static and independent token-level distribution over the whole candidate set, ignoring the specificity of certain a sample, they may wrongly encourage a token that is rare in the candidate set but occurs many times in the sample (\eg proper nouns).
Besides, \citet{yi-etal-2020-improving} demonstrates that the tokens with high IDF are not always indicative of semantic similarity due to the co-occurrences.
Second, attention from different representation layers of BERT has been proven with different semantic and reasoning abilities~\cite{van2019does} to encode its input. For example, the shallow layer is used for named entity labeling, the middle layer for coreference resolution, and the deep layer for relation classification.
Thereby, layer-wise contextualized embeddings and attention of BERT can be engaged to capture different relationships between tokens to evaluate the word-level relevance reasonably and hierarchically, \ie from superficial relationships to complicated ones.

Formally, we firstly obtain the dynamic contextualized embeddings as follows:
\begin{equation}
    \left\{  a_{mn}^{l}, f\left(\hat{x}_{m}\right)^{l}, f\left(c_{n}\right)^{l}  \right\}_{l=1}^{L} = \text{BERT}(  [ \hat{X}, C  ] )
\end{equation}
where $a_{mn}^{l}$ represents the maximum of normalized attention scores among all heads at the $l$-th layer of BERT between the $m$-th token in the candidate and the $n$-th token in the context, while $f\left(\hat{x}_{m}\right)^{l}$ and $f\left(c_{n}\right)^{l}$ denote the contextualized embeddings of corresponding tokens at the same layer and $[ \hat{X}, C ]$ means the concatenation of the candidate and the context.
Then, our QRel$_{LRM}$ is computed as the precision-based cosine similarity of tokens from the candidate and the context, where each token in the candidate is matched to the token in the context with an aggregate function and dynamic weighting.
After that, our QRel$_{LRM}$ merges the layer-wise relevance score with power means~\cite{ruckle2018concatenated}, which is an effective generalization of pooling techniques for multi-level information.
\begin{align}
    \text{QRel}_{LRM} &= \sqrt[\uproot{20}\scriptstyle p]{ \frac{1}{L} \sum_{l=1}^{L} \text{Prec}_{l\text{-th}}^{p} } \\
    \text{Prec}_{l\text{-th}} &= \frac{1}{M} \sum_{m}^{M}{\underset{c_{n} \in C}{agg} \left( a_{mn}^{l} f\left(\hat{x}_{m}\right)^{l} \odot f\left(c_{n}\right)^{l} \right)} \label{eq:qrellr-lth}
\end{align}
where $p$, $\odot$ and $agg(\cdot)$ represent the exponent of power means, the cosine similarity and an aggregate function, respectively.
Empirically, we set $p=1$ and the $agg(\cdot)$ as a $max$ function.
    
In practice, we observe that the layer-wise scores are in a more limited range (around $-0.1\sim0.4$), potentially because of the learned geometry of contextualized embeddings from language models.
Following the widely adopted solutions~\cite{zhang2019bertscore, hessel2021clipscore}, we linearly rescale QRel$_{LRM}$ with its lower bound $b_{LRM}$ as a baseline to put it between 0 and 1.
Empirically, we compute the $b_{LRM}$ by averaging QRel$_{LRM}$ on the random $\langle$ \emph{candidate}, \emph{context} $\rangle$ pairs on the corresponding dataset.
Notice that the min-max normalization has the same effect as this linear rescaling.

\begin{equation}
    \text{QRel}_{LRM} = \frac{\text{QRel}_{LRM} - b_{LRM}}{1 - b_{LRM}}
\end{equation}

\begin{figure}[!t]
\begin{tcolorbox}
    \small
    \textbf{Context.} Jack drove his car to the bazaar to purchase milk and honey for his large family.
    
    \textbf{Reference (0.905).} Where \textcolor{dgreen}{did} Jack buy \textcolor{dgreen}{his} \textcolor{dgreen}{milk and honey}?
    
    \textbf{Entity swap (0.816).} Where did Jack buy his \textcolor{dred}{car}?
    
    \textbf{Pronoun swap (0.847).} Where did Jack buy \textcolor{dred}{your} milk and honey?
    
    \textbf{Sentence negation (0.803).} Where \textcolor{dred}{didn't} Jack buy his milk and honey?
\end{tcolorbox}
\caption{Three unanswerable example questions constructed by perturbing only the individual words.
Their QRel$_{LRM}$ scores (marked in the round brackets) do not reflect the factual inconsistency ideally.
}
\label{fig:qrellr-pitfalls}
\vspace{-5mm}
\end{figure}

\subsection{Global Relevance Generation}
Although QRel$_{LRM}$ can measure the word-level relevance of QG, candidates that contain a group of semantically similar tokens to the context, but ungrammatical or incoherent, can also receive a relatively high score.
In this case, QRel$_{LRM}$ fails to ideally penalize the factual inconsistency arising from the individual words and capture multiple evidences in the context.
Figure~\ref{fig:qrellr-pitfalls} shows some pitfalls of QRel$_{LRM}$.
To mitigate this problem and achieve a robust measure of the global relevance, we further devise QRel$_{GRG}$ based on the prompt of causal language models (CLMs) such as GPT2.

Prompt-based learning maximizes the generalization capability of language models and is becoming a new paradigm in natural language processing~\cite{liu2021pre}.
In this paper, we formulate our QRel$_{GRG}$ as the confidence gain by \textit{comparing the likelihood of generating the context with or without the candidate as a prompt}.
Our QRel$_{GRG}$ appropriately encourages the candidate that is highly relevant to the context because a question inconsistent with the context is pretty likely to make a limited or even negative difference to the unidirectional generation.
Based on the confidence difference caused by the candidate, QRel$_{GRG}$ measures the overall relevance between the candidate and all the possible evidences in the context.

More precisely, causal language modeling, also known as autoregressive language modeling, is a classic probabilistic density estimation problem.
Given an input sequence $S = \langle s_1, \cdots, s_t, \cdots, s_T \rangle$, its joint distribution $p(S)$ or $p(s_{1:T})$ can be decomposed as:
\begin{equation}
    p(S) =  \prod_{t=1}^{T}{p\left( s_t \mid s_{0:t-1} \right)}
\end{equation}
where $s_0$ is a special token indicating the begin of sequence and $p\left( s_t \mid s_{0:t-1} \right)$ represents the tractable conditional probabilities $p\left( s_t \mid s_0, \cdots, s_{t-1} \right)$.
Abbreviating $p\left( s_t \mid s_{0:t-1} \right)$ as $p_{s_t}$, we feed the $C$ and $[ \hat{X}, C ]$ into the GPT2 successively to obtain the conditional probability of every token in the context as follows:
\begin{align}
    \left\{  p_{c_n} \right\}_{n=1}^{N} &= \text{GPT2}\left( C \right) \\
    \left\{  p^{\prime}_{\hat{x}_m} \right\}_{m=1}^{M}, \left\{  p^{\prime}_{c_n} \right\}_{n=1}^{N} &= \text{GPT2}( [ \hat{X}, C ] )
\end{align}
After that, the baseline confidence $\text{Conf}_{base}$ and prompted confidence $\text{Conf}_{prompt}$ are computed as:
$\text{Conf}_{base} = \sum_{n=1}^{N}{ \log p_{c_n} }$ and $\text{Conf}_{prompt} = \sum_{n=1}^{N}{ \log p^{\prime}_{c_n} }$, respectively.
Finally, our QRel$_{GRG}$ is quantified as the gain ratio of the confidence caused by the candidate.
\begin{equation}
    \text{QRel}_{GRG} = \max\left\{ \frac{\text{Conf}_{prompt} - \text{Conf}_{base}}{\vert \text{Conf}_{base} \vert}, 0 \right\} \label{eq:qrelga}
\end{equation}
For the same reason as QRel$_{LRM}$, we rescale the QRel$_{GRG}$ with $b_{GRG}$ to increase the readability of this score and do not affect its ranking ability or correlation with human judgments.

\subsection{Reference-augmented QRelScore}
QRelScore can additionally be extended to incorporate references if they are available.
Specifically, given a set of human references $R$, Ref-QRelScore is computed as the arithmetic mean
of QRelScore between the candidate and context, and maximal QRelScore between the candidate and reference.
\begin{equation}
\begin{split}
    & \text{Ref-QRelScore}(\hat{X}, C, R) = \\
    & \frac{1}{2}(\text{QRelScore}(\hat{X}, C) + \max_{r \in R}\text{QRelScore}(\hat{X}, r))
\end{split}
\end{equation}

\section{Experiments}
\label{sec:experiments}
\noindent
\textbf{Datasets.} \
We employ two widely-used QG datasets to validate our QRelScore, including SQuADv1~\cite{rajpurkar-etal-2016-squad} and HotpotQA~\cite{yang-etal-2018-hotpotqa}.
We re-divide the SQuADv1 dataset into train/dev/test splits following \citet{zhou2017neural}.
For the HotpotQA dataset, we utilize the official train/dev/test splits.

\noindent
\textbf{Candidate questions.} \
We obtain two candidate sets of shallow questions (\ie factoid questions) respectively from NQG++~\cite{zhou2017neural} and BART-QG~\cite{lewis-etal-2020-bart} on the SQuADv1 dataset, and another two candidate sets of more complicated questions that require reasoning over multiple pieces of information respectively from DP-Graph~\cite{pan-etal-2020-semantic} and DCQG~\cite{cheng-etal-2021-guiding} on the HotpotQA dataset.

\noindent
\textbf{Implementation details.} \
Our QRel$_{LRM}$ and QRel$_{GRG}$ are implemented by BERT-base (12-layer) and OpenAI GPT2 (12-layer) English models, respectively.
The contextualized embeddings and attention scores of BERT-base and generation likelihood of GPT2 are extracted by the HuggingFace Transformers package~\cite{wolf-etal-2020-transformers}.
In case of the input exceeding the maximum length acceptable to the language models (\ie 512 and 1024 tokens for BERT and GPT2, respectively), we first cut the long context into several text chunks with maximum acceptable length.
Then they are fed into the model one by one, along with the candidate question.
After that, the final score is calculated by averaging the relevance scores across all chunks.

\noindent
\textbf{Baselines.} \
We verify the effectiveness of our QRelScore by comparing it to the following three types of evaluation metrics.
Firstly, we choose traditional n-gram matching based metrics including BLEU-4~\cite{papineni-etal-2002-bleu}, ROUGE-L~\cite{lin-2004-rouge} and METEOR~\cite{banerjee-lavie-2005-meteor}.
Furthermore, we also extend more recent reference-based methods as baselines such as Q-BLEU~\cite{nema-khapra-2018-towards}, BERTScore~\cite{zhang2019bertscore}, Moverscore~\cite{zhao-etal-2019-moverscore}, BLEURT~\cite{sellam-etal-2020-bleurt} and COMET~\cite{rei-etal-2020-comet}.
The last two baselines are supervised metrics optimized by the regression and ranking objective, respectively.
In addition, we construct two reference-free baselines by replacing the reference input of Q-BLEU and BERTScore with the corresponding context, which is denoted as Q-BLEU$_{free}$ and BERTScore$_{free}$, respectively.
At last, we adopt two state-of-the-art reference-free factuality evaluation metrics in the abstractive summarization task as our baselines, including the embedding-based consistency dimension of CTC~\cite{deng-etal-2021-compression} and the faithfulness dimension of BARTScore~\cite{yuan2021bartscore}.

\noindent
\textbf{Human annotation.} \
Because the examined QG models do not release corresponding human evaluation results on the quality of their generated questions,
we first evaluate the quality of the generated candidate via voluntary human evaluation.
Following the human criteria of QG elaborated by \citet{rus-etal-2010-first} and \citet{nema-khapra-2018-towards}, we annotate each sample in terms of grammaticality, answerability, and relevance.
Specifically, we ask five annotators to rate the quality of 800 $\langle$ \emph{passage}, \emph{question}, \emph{answer} $\rangle$ candidates from the four models, including NQG++, BART-QG, DP-Graph and DCQG , with 200 candidates per model.
All the samples are randomly shuffled and anonymized.
The annotators are informed of the detailed annotation instruction with clear scoring examples and evaluate the grammaticality and answerability on a 3-point Likert scale (1 for ``unacceptable'' and 3 for ``excellent''), while the relevance dimension with a binary rating (1 for ``relevant'' and 2 for ``irrelevant'').
\emph{Please refer to Appendix~\ref{sec:annotation-details} for more details about human annotation.}

\subsection{Main Results}
\noindent
\textbf{Human-human correlation.} \
The inter-annotator agreement for the four candidate sets is 82.43\%, 85.39\%, 85.81\%, and 87.25\%, respectively. 
We use the average of five corresponding annotator ratings as the final human judgment for a specific dimension of a given candidate question.

\begin{table}[!t]
    \centering
     \resizebox{1.0\columnwidth}{!}{
        \begin{tabular}{c|ccc|ccc|ccc}
            \toprule
            \multicolumn{1}{c}{\multirow{2}{*}{\textbf{Metrics}}}& \centerthreeg{\textbf{Grammaticality}}& \centerthreeb{\textbf{Answerability}}& \centerthreer{\textbf{Relevance}} \\
            \multicolumn{1}{c}{}& \centeroneg{\textbf{$r$}}& \centeroneg{\textbf{$\rho$}}& \centeroneg{\textbf{$\tau$}}& \centeroneb{\textbf{$r$}}& \centeroneb{\textbf{$\rho$}}& \centeroneb{\textbf{$\tau$}}& \centeroner{\textbf{$r$}}& \centeroner{\textbf{$\rho$}}& \centeroner{\textbf{$\tau$}} \\
            \midrule
            \midrule
                BLEU-4& 0.154& 0.145& 0.144& 0.198& 0.178& 0.139& 0.134& 0.111& 0.102 \\
                ROUGE-L& 0.180& 0.185& 0.184& 0.217& 0.207& 0.162& 0.164& 0.136& 0.122 \\
                METEOR& 0.195& 0.183& 0.185& 0.245& 0.225& 0.169& 0.176& 0.148& 0.131 \\
                \hline
                Q-BLEU& 0.313& 0.304& 0.301& 0.349& 0.322& 0.258& 0.285& 0.267& 0.226 \\
                BERTScore& 0.338& 0.350& 0.336& 0.389& 0.368& 0.272& 0.305& 0.287& 0.254 \\
                MoverScore& 0.370& 0.362& 0.359& 0.396& 0.377& 0.299& 0.320& 0.305& 0.255 \\ 
                BLEURT& 0.397& 0.398& 0.366& 0.414& 0.385& 0.322& 0.342& 0.323& 0.282 \\
                COMET& 0.442& 0.435& 0.431& 0.461& 0.444& 0.353& 0.381& 0.370& 0.306 \\
                \hline
                Q-BLEU$_{free}$& 0.394& 0.387& 0.354& 0.398& 0.371& 0.318& 0.324& 0.311& 0.263 \\
                BERTScore$_{free}$& 0.434& 0.386& 0.399& 0.428& 0.412& 0.332& 0.372& 0.341& 0.285 \\
                CTC& 0.446& 0.441& 0.437& 0.463& 0.445& 0.356& 0.383& 0.372& 0.309 \\
                BARTScore& 0.453& 0.444& 0.441& 0.473& 0.453& 0.360& 0.389& 0.379& 0.314 \\
                \hline
                \underline{QRelScore}& \underline{0.496}& \underline{0.489}& \underline{0.485}& \underline{0.511}& \underline{0.493}& \underline{0.393}& \underline{0.428}& \underline{0.418}& \underline{0.345} \\
                \textbf{Ref-QRelScore}& \textbf{0.516}& \textbf{0.508}& \textbf{0.504}& \textbf{0.528}& \textbf{0.508}& \textbf{0.407}& \textbf{0.441}& \textbf{0.435}& \textbf{0.359} \\
            \bottomrule
        \end{tabular}
     }
    \caption{Segment-level correlation in Pearson’s $r$, Spearman’s $\rho$, and
    Kendall’s $\tau$ with human judgments on the SQuADv1 dataset.
    The best and second-best results are \textbf{bold} and \underline{underlined}, respectively.}
    \label{tab:segment-correlation}
    \vspace{-5mm}
\end{table}

\noindent
\textbf{Human-score correlation.} \
Table~\ref{tab:segment-correlation} presents segment-level correlation to human judgments on SQuADv1.
We observe that our QRelScore consistently outperforms all the baselines in terms of answerability and relevance, which indicates the effectiveness of incorporating context-aware relevance into the evaluation of QG.
In addition, the better grammaticality correlations attribute to the autoregressive language model in our QRelScore, which measures the naturalness and fluency of the candidate more accurately by considering the word-by-word human language properties.
When incorporating the available human reference into our metric, Ref-QRelScore achieves an even higher correlation with human judgments.

\begin{figure}[t]
	\centering

	\begin{subfigure}{0.49\columnwidth}
    \includegraphics[width=\textwidth]{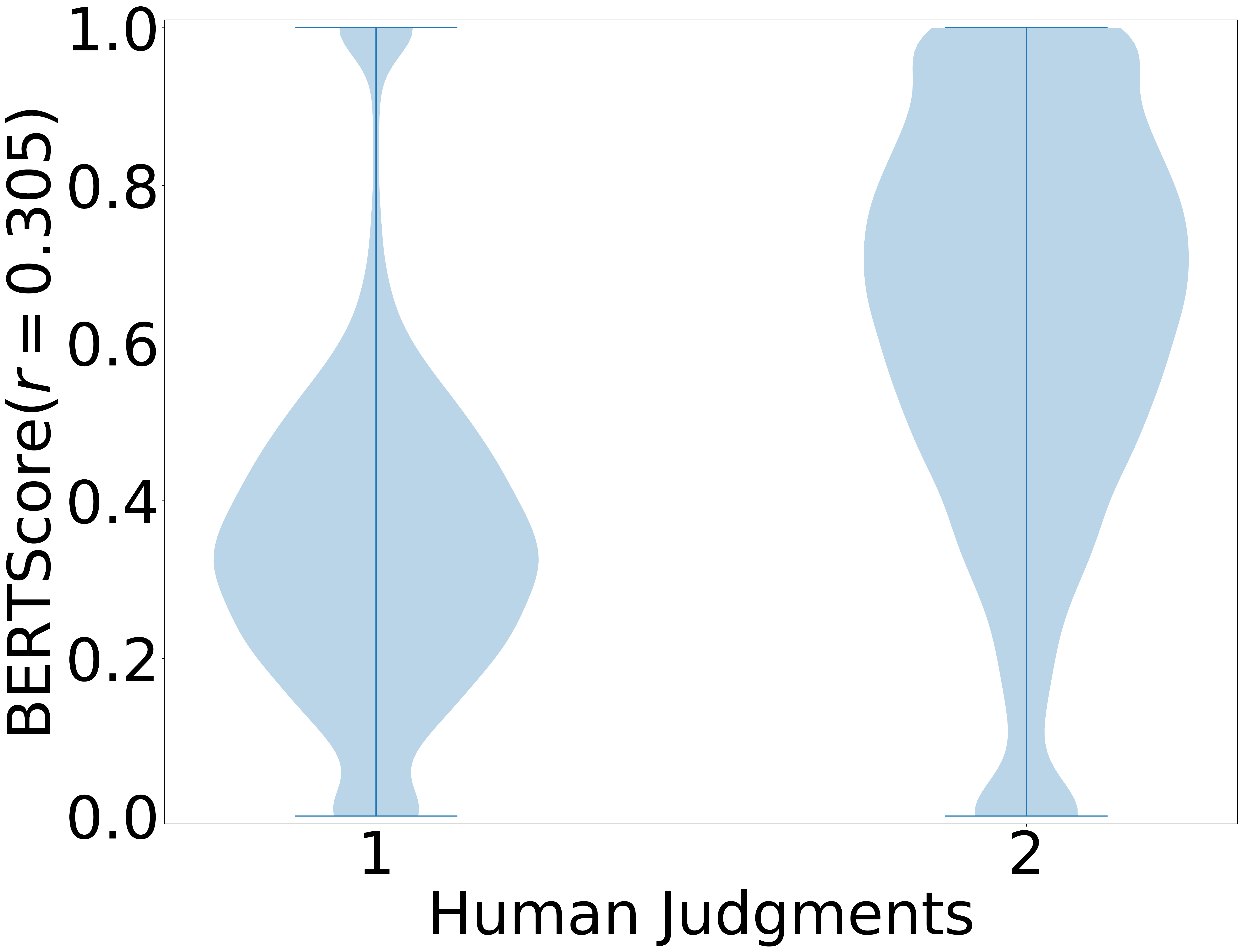}
	\end{subfigure}
	\hfill
	\begin{subfigure}{0.49\columnwidth}
    \includegraphics[width=\textwidth]{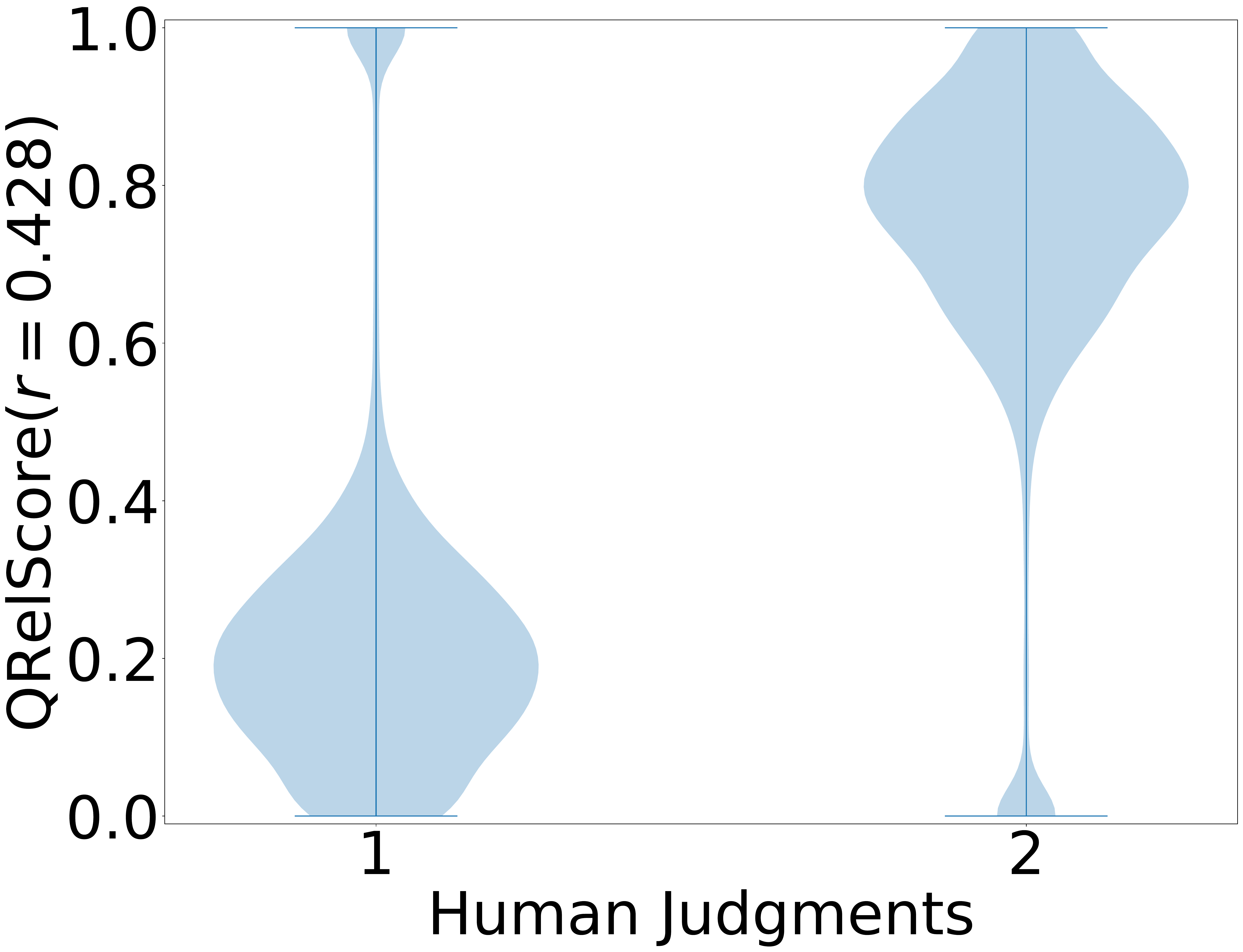}
	\end{subfigure}

    \caption{Score distributions of BERTScore and QRelScore under different relevance ratings (\ie 1 for ``relevant'' and 2 for ``irrelevant'') of human judgments.}
     \label{fig:score-distribution}
     \vspace{-5mm}
\end{figure}

In Figure~\ref{fig:score-distribution}, we take a closer look at the correlation results by the distribution of scores w.r.t. the human judgments.
Results reveal that previous metrics such as BERTScore can correctly assign lower scores to the candidates of low quality (``1'' relevance rating), but it performs poorly in the candidates of high quality (``2'' relevance rating).
This finding agrees with the motivation of our work, namely that lacking a deep understanding of the context-aware relevance may lead to a wrong penalization to the legitimate and reasonable candidate.
Importantly, our QRelScore can clearly distinguish the candidates with different qualities.
\emph{Please refer to Appendix~\ref{sec:more-experiments} for more experimental results.}

\subsection{Ablation Analysis}
We conduct our ablation experiments and summarize the quantitative results in Table~\ref{tab:ablation} from the following three aspects.

First, whether QRel$_{LRM}$ or QRel$_{GRG}$ alone is sufficient to evaluate the relevance of QG?
The first two baselines compute the relevance score by QRel$_{LRM}$ or QRel$_{GRG}$ only, denoted as ``QRel$_{LRM}$ ($M_1$)'' and ``QRel$_{GRG}$ ($M_8$)'', respectively.
As shown in the table, both QRel$_{LRM}$ and QRel$_{GRG}$ make significant contributions to the final performance.
For example, both $M_1$ and $M_8$ also outperform previous metrics (in Table~\ref{tab:segment-correlation}) in terms of three dimensions.
This result attributes to the incorporation of the word- and sentence-level relevance into the evaluation metrics.

\begin{table}[!t]
    \centering
     \resizebox{1.0\columnwidth}{!}{
        \begin{tabular}{c|c|ccc|ccc|ccc}
            \toprule
            \multicolumn{1}{c}{\multirow{2}{*}{\textbf{Name}}}&              \multicolumn{1}{c}{\multirow{2}{*}{\textbf{Metrics}}}& \centerthreeg{\textbf{Grammaticality}}& \centerthreeb{\textbf{Answerability}}& \centerthreer{\textbf{Relevance}} \\
            \multicolumn{1}{c}{}& \multicolumn{1}{c}{}& \centeroneg{\textbf{$r$}}& \centeroneg{\textbf{$\rho$}}& \centeroneg{\textbf{$\tau$}}& \centeroneb{\textbf{$r$}}& \centeroneb{\textbf{$\rho$}}& \centeroneb{\textbf{$\tau$}}& \centeroner{\textbf{$r$}}& \centeroner{\textbf{$\rho$}}& \centeroner{\textbf{$\tau$}} \\
            \midrule
            \midrule

                $M_1$& \textbf{QRel$_{LRM}$}& \textbf{0.478}& \textbf{0.471}& \textbf{0.467}& \textbf{0.494}& \textbf{0.477}& \textbf{0.380}& \textbf{0.412}& \textbf{0.402}& \textbf{0.332} \\
                $M_2$& w/ first& 0.370& 0.364& 0.362& 0.394& 0.376& 0.300& 0.319& 0.304& 0.256 \\
                $M_3$& w/ middle& 0.406& 0.397& 0.395& 0.430& 0.410& 0.326& 0.349& 0.337& 0.281 \\
                $M_4$& w/ last& 0.425& 0.417& 0.413& 0.446& 0.425& 0.340& 0.366& 0.355& 0.296 \\
                $M_5$& w/ specific& 0.442& 0.436& 0.431& 0.463& 0.443& 0.352& 0.380& 0.370& 0.309 \\
                $M_6$& w/ average& 0.444& 0.437& 0.431& 0.462& 0.441& 0.354& 0.381& 0.370& 0.309 \\
                $M_7$& w/ mover& 0.464& 0.456& 0.448& 0.478& 0.457& 0.368& 0.395& 0.386& 0.321 \\
                \hline
                $M_8$& \textbf{QRel$_{GRG}$}& \textbf{0.464}& \textbf{0.451}& \textbf{0.450}& \textbf{0.478}& \textbf{0.458}& \textbf{0.367}& \textbf{0.397}& \textbf{0.386}& \textbf{0.320} \\
                $M_9$& w/ absolute& 0.390& 0.381& 0.378& 0.412& 0.394& 0.314& 0.334& 0.323& 0.268 \\
                \hline
                & \textbf{QRelScore}& \textbf{0.496}& \textbf{0.489}& \textbf{0.485}& \textbf{0.511}& \textbf{0.493}& \textbf{0.393}& \textbf{0.428}& \textbf{0.418}& \textbf{0.345} \\
            \bottomrule
        \end{tabular}
     }
    \caption{Segment-level correlation in terms of Pearson’s $r$, Spearman’s $\rho$, and
    Kendall’s $\tau$ with human judgments on the SQuADv1 dataset.
    In the table, the \emph{upper} part is for the ablation analysis of QRel$_{LRM}$, while the \emph{lower} part is for QRel$_{GRG}$.
    The best results are highlighted in \textbf{bold}.}
    \label{tab:ablation}
    \vspace{-5mm}
\end{table}

Second, for QRel$_{LRM}$, what is the impact of the number of levels of attention scores and the way it aggregates the semantically similar tokens in the context for a token in the candidate?
On the one hand, ``QRel$_{LRM}$ w/ first ($M_2$)'', ``QRel$_{LRM}$ w/ middle ($M_3$)'', ``QRel$_{LRM}$ w/ last ($M_4$)'' and ``QRel$_{LRM}$ w/ specific ($M_5$)'' use the first four layers (0, 1, 2, 3), the middle four layers (4, 5, 6, 7), the last four layers (8, 9, 10, 11) and specific four layers (0, 3, 7, 11) of BERT attention, respectively.
The experimental results reveal that $M_2$, $M_3$, $M_4$ and $M_5$ degrade the performance w.r.t. $M_1$ in three dimensions, demonstrating the attention at different layers plays an irreplaceable role in final results.
Among them, $M_5$ achieves the best correlation, which shows the necessity of evaluating the relevance in a progressive manner, that is, from the shallow layer to the deep one.
On the other hand, ``QRel$_{LRM}$ w/ average ($M_6$)'' and ``QRel$_{LRM}$ w/ mover ($M_7$)'' substitute the $max$ operation in Eq.~\ref{eq:qrellr-lth} with an $avg$ function and a $sum$ function weighted by the probability transitive matrix, which is obtained by optimizing earth mover's distance (EMD)~\cite{rubner1998metric} from the candidate to the context on each layer.
According to the results in Table~\ref{tab:ablation}, $M_6$ and $M_7$ show worse correlation than $M_1$, verifying that the averaging aggregation and optimal transportation optimization result in a biased relevance evaluation.
A possible reason is that they fail to capture the token-wise specificity because average-based aggregation weakens the effects of irrelevant tokens and hinders the discriminative ability of the metrics.

Third, for QRel$_{GRG}$, is the absolute confidence gain a better relevance criterion?
``QRel$_{GRG}$ w/ absolute ($M_9$)'' computes the global relevance by directly subtracting the $\text{Conf}_{base}$ from $\text{Conf}_{prompt}$ in Eq.~\ref{eq:qrelga}.
From the results in Table~\ref{tab:ablation}, $M_9$ degrades performance w.r.t. $M_8$ significantly, showing that the absolute confidence gain is not a proper measurement for sentence-level relevance since it takes account of the factors unrelated to the generation quality, such as the length of the candidate and the domain effects of pre-trained language models.

\begin{figure}[t]
	\centering

	\begin{subfigure}{0.49\columnwidth}
    \includegraphics[width=\textwidth]{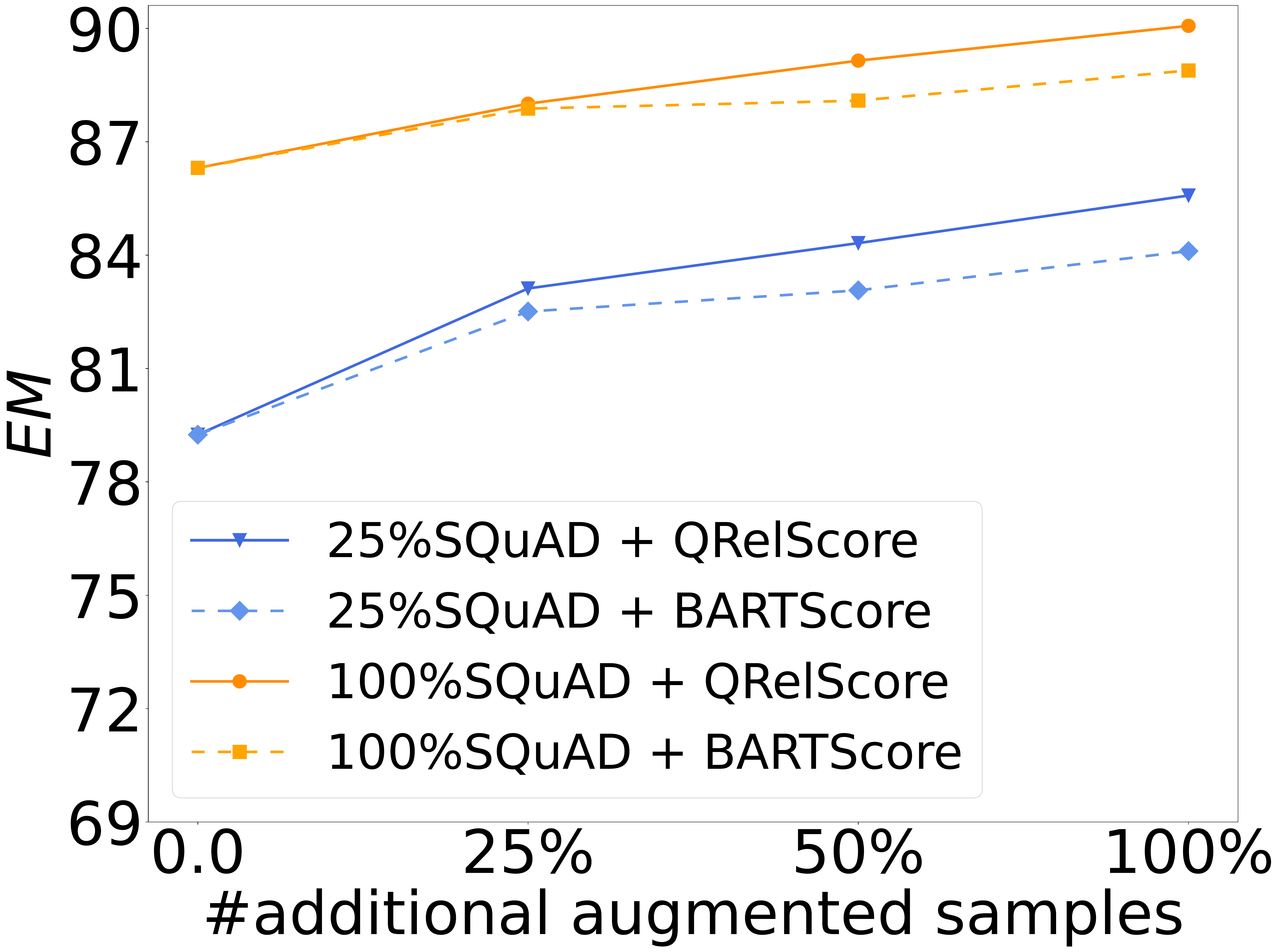}
	\end{subfigure}
	\hfill
	\begin{subfigure}{0.49\columnwidth}
    \includegraphics[width=\textwidth]{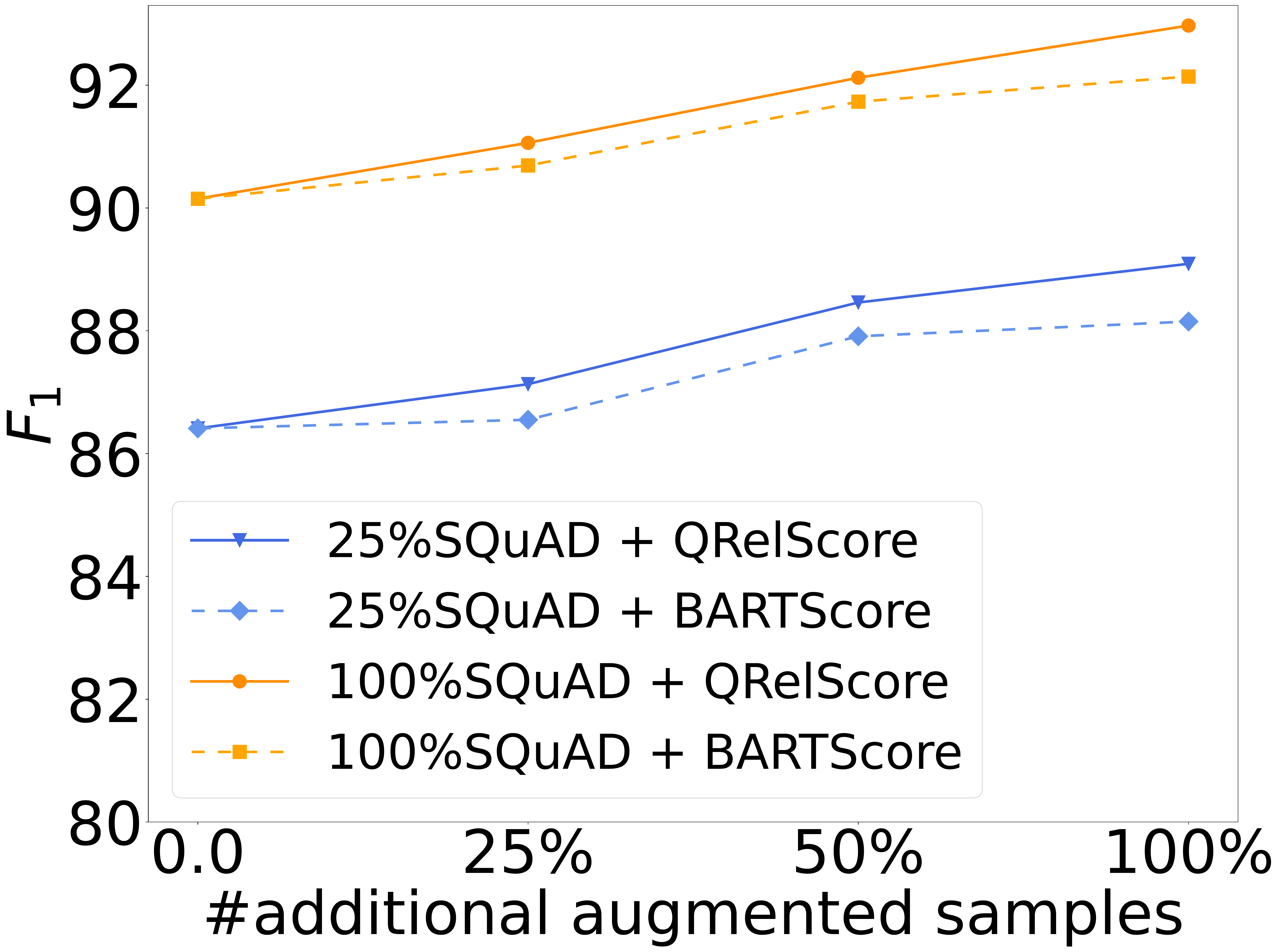}
	\end{subfigure}

    \caption{Performance of the DistilBERT-based QA system on the SQuADv1 dataset, augmented with the data generated by different QG rewards.}
     \label{fig:augmented-qa}
     \vspace{-5mm}
\end{figure}

\subsection{Evaluate QRelScore Rewards for QG with Reinforcement Learning (RL)}
To further demonstrate the superiority of our QRelScore, we employ the QRelScore as a reward to optimize the RL-based QG and evaluate the quality of generated questions with a QA system.
Specifically, we embed BART-QG into a self-critical sequence training (SCST) framework~\cite{rennie2017self} and compute the reward using our QRelScore.
After that, the whole pipeline is optimized on the aforementioned train split of the SQuADv1 dataset to generate questions conditioned on the context and answer.
During inference, we do not feed unseen paragraphs (\ie the paragraphs in the dev or test split) into the model, and just make the BART-QG model generate diverse questions for the existing paragraphs in the SQuADv1 training set by keeping all beam search (size$=8$) outputs for each sample.
Furthermore, we filter out the obviously low-quality questions if their word counts are not between $6\sim30$, or the answers directly appear in the questions.
Finally, we randomly sample 90,000 QA pairs and augment the SQuADv1 training dataset with them.
As a comparison, we design a baseline by employing BARTScore as the RL reward, which is the most competitive metric in Table~\ref{tab:segment-correlation}.

A DistilBERT-based~\cite{sanh2019distilbert} QA model is trained on this augmented dataset to evaluate the quality of generated questions.
The comparisons of QA performance under high-resource setting (using the whole training set of SQuADv1) and low-resource setting (using 25\% of data sampled from SQuADv1) are illustrated in Figure~\ref{fig:augmented-qa}.
We can observe that BART-QG with QRelScore as the reward achieves better performance than BARTScore under both settings.
As more and more of our generated data is added to the training set, the QA performance gets better and better and reaches a 4.36\% / 3.13\% achievement of $EM$/$F_1$ when the number of additional augmented samples reaches the size of the SQuADv1 training set.

\begin{table}[!t]
    \centering
     \resizebox{1.0\columnwidth}{!}{
        \begin{tabular}{c|c|cc}
            \toprule
            \textbf{Type}& \textbf{Method}& \textbf{SQuADv1}& \textbf{HotpotQA} \\
            \midrule
            \midrule
              \multirow{3}{2cm}{\centering Supervised models}& DecAtt& 0.791& 0.641 \\
              & DIIN& 0.852& 0.718\\
              & \textbf{BERT}& \textbf{0.943}& \textbf{0.801} \\
              \hline
              \multirow{10}{*}{Metrics}& BLEU-4& 0.698& 0.527 \\
                & ROUGE-L& 0.703& 0.533 \\
                & METEOR& 0.712& 0.542 \\
                \cline{2-4}
                & Q-BLEU& 0.733& 0.566 \\
                & BERTScore& 0.740& 0.575 \\
                & MoverScore& 0.751& 0.588 \\
                & BLEURT& 0.773& 0.612 \\
                & COMET& 0.798& 0.643 \\
                \cline{2-4}
                & Q-BLEU$_{free}$& 0.767& 0.606 \\
                & BERTScore$_{free}$& 0.788& 0.630 \\
                & CTC& 0.808& 0.653 \\
                & BARTScore& 0.815& 0.661 \\
                \cline{2-4}
                & \textbf{QRelScore}& \textbf{0.844}& \textbf{0.690} \\
            \bottomrule
        \end{tabular}
     }
    \caption{Area under the ROC curve (AUC) of classifying adversarial samples on SQuADv1 and HotpotQA datasets.
    The best results are highlighted in \textbf{bold}.}
    \label{tab:adversarial-results}
    \vspace{-5mm}
\end{table}

\subsection{Robustness Analysis}
A competent evaluation metric can not only distinguish between good and bad systems but also help analyze the samples.
Therefore, we test the robustness of QRelScore by detecting adversarial samples.
Specifically, inspired by the major types of relevance and factuality errors in the text generation~\cite{goyal-durrett-2020-evaluating, chen-etal-2021-factuality-checkers, pagnoni-etal-2021-understanding}, we construct the positive samples by paraphrasing transformation.
In contrast, negative samples are generated by the swapping and negation perturbations, including entity, pronoun swapping, and sentence negation.
We generate 10,000 positive and 10,000 negative samples using the randomly chosen samples from the SQuADv1 and HotpotQA dev set as the positive anchors and employ the QRelScore to classify them based on the relevance scores.
In addition to existing automatic metrics, we also fine-tune three supervised baselines ten times in 5-fold cross-validation, including DecAtt~\cite{parikh-etal-2016-decomposable}, DIIN~\cite{gong2018natural} and BERT~\cite{devlin-etal-2019-bert}.
\emph{Please refer to Appendix~\ref{sec:adversarial-details} for the details on the adversarial examples.}

\begin{table}[!t]
    \centering
     \resizebox{1.0\columnwidth}{!}{
        \begin{tabular}{ll}
            \toprule
            \textbf{Error}& \textbf{Example} \\
            \midrule
            \multirow{8}{2cm}{\centering Out of Vocabulary}& \textbf{Context:} $[\ldots]$ The 2012 Washington \\
            & State Cougars football team was \\
            & coached by first-year head coach \\
            & Mike Leach. $[\ldots]$ \\
            & \textbf{Candidate:} Where does UNK UNK \\
            & currently coach at? \\
            & \textbf{Human:} (0.600, 0.667 0.800) \\
            & \textbf{QRelScore:} 0.198 \\
            \hline
            \multirow{5}{*}{Confusion}& \textbf{Context:} $[\ldots]$ Jacob put the marbles \\
            & in the box and the bowl on the table. $[\ldots]$ \\
            & \textbf{Candidate:} Where did he put the marbles? \\
            & \textbf{Human:} (1.000, 0.333, 0.867) \\
            & \textbf{QRelScore:} 0.821 \\
            \hline            
            \multirow{8}{2cm}{\centering Domain-specific Knowledge}& \textbf{Context:} $[\ldots]$ Denver continued to \\
            & pound away as RB Cecil Sapp got a \\
            & 4-yard TD run, while kicker Jason \\
            & Elam got a 23-yard field goal. $[\ldots]$ \\
            & \textbf{Candidate:} Which position scored \\
            & the shortest touchdown of the game? \\
            & \textbf{Human:} (1.000, 1.000, 0.933) \\
            & \textbf{QRelScore:} 0.206 \\
            
            \bottomrule
        \end{tabular}
     }
    \caption{Three typical types of errors found in the samples which received significant differences between the QRelScore and human judgments.
    }
    \label{tab:error-analysis}
    \vspace{-5mm}
\end{table}

Table~\ref{tab:adversarial-results} reports the area under the ROC curve (AUC) for QRelScore and other baselines.
As shown in the table, compared to the supervised BERT classifier, most of the metrics degrade performance significantly.
However, some metrics, including our QRelScore, outperform a relatively weak model (\ie DecAtt).
This suggests that these metrics have a certain level of ability to detect adversarial samples.
Among all the metrics, our QRelScore achieves the best results and shows the slightest performance drop on both datasets, showing more robustness than the other metrics.

\subsection{Error Analysis}
We analyze cases where the QRelScore substantially differs from human judgments.
In Table~\ref{tab:error-analysis}, these error cases can be categorized into one of three types:
(1) Out of vocabulary errors, often induced by unknown tokens in the candidates, (2) Confusion errors, the scope of coordination may be interpreted differently and thus lead to a syntactic ambiguity, \eg the marbles were either put both in the box and in the bowl that was on the table, or the marbles were put in the box and the bowl was put on the table, and (3) Knowledge errors, where the candidates are further inferences based on the commonsense knowledge or domain-specific knowledge, \eg in showing cases, both running back (RB) and kicker (K) are the positions of a player on an American or Canadian football team.
These errors reveal the limitations of our QRelScore and give us directions for future improvement by engaging language models with a larger capacity.

\section{Related Work}
\label{sec:related}

\noindent
\textbf{Aspect-specific evaluation.} \
Some words measured the semantic similarity between the text by leveraging the static word representations~\cite{lo-2017-meant, clark-etal-2019-sentence}, contextualized BERT embedding~\cite{zhang2019bertscore, zhao-etal-2019-moverscore}, or fine-tuning on human-rated quality scores~\cite{sellam-etal-2020-bleurt, rei-etal-2020-comet}.
In an unified formulation, the recent-emerged approaches devised a family of metrics to evaluate different natural language generation (NLG) tasks.
For example, CTC~\cite{deng-etal-2021-compression} evaluated the information alignment between the text from three aspects of NLG, including compression, transduction, and creation, while BARTScore~\cite{yuan2021bartscore} presented multiple evaluation aspects based on different generation directions.
Although it was similar to the QRel$_{GRG}$ of our QRelScore in some way, it employed the absolute likelihood of generation and required extra fine-tuning to reduce the domain effects.

\noindent
\textbf{Relevance and factual consistency.} \
Relevance is widely explored in the response coherence of dialogue system~\cite{huang-etal-2020-grade} and factuality of document summarization~\cite{gabriel-etal-2021-go} besides question generation.
\citet{kryscinski-etal-2020-evaluating} proposed a weakly-supervised approach for verifying the factual consistency of a generated summary.
On a broader scale, \citet{maynez-etal-2020-faithfulness} conducted an extensive human evaluation of several summarization systems and analyzed the types of factual hallucinations they produced.
More recently, MARS~\cite{liu-etal-2021-language} was proposed to evaluate relevance by augmented references.
We considered lessons of context-awareness from these works while designing the QRelScore.

\section{Conclusion}
\label{sec:conclusion}

Existing evaluation metrics for question generation are still reference-based and ignore the crucial input context of generation, lacking a deep understanding of the relevance between the generated questions and input context.
To address these issues, we propose QRelScore, which measures the word- and sentence-level relevance without additional training and human supervision by the off-the-shelf language models.
Extensive experiments demonstrate that our QRelScore achieves start-of-the-art correlation with human judgments and makes up for the shortcomings of existing reference-based metrics.



\bibliography{anthology, custom}

\appendix
\clearpage

\section{Annotation Details}
\label{sec:annotation-details}

\begin{figure*}[!t]
    \includegraphics[width=\textwidth]{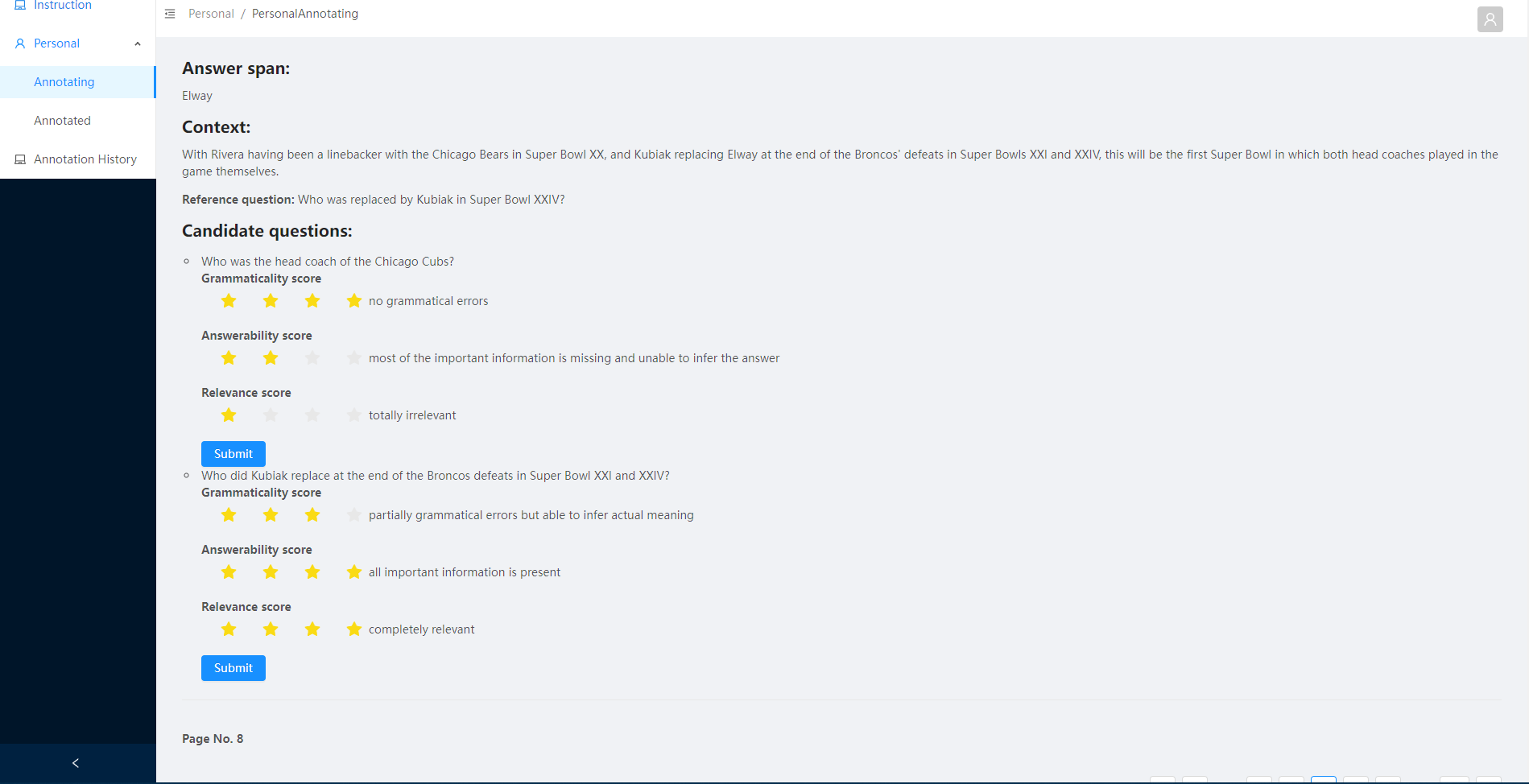}
    \caption{A screenshot of our human annotation process.}
    \label{fig:software}
\end{figure*}

A total of five annotators participated in our study.
The annotators were Computer Science graduates competent in English and kindly offered their help as volunteers without being compensated in any form.
All the samples from the three examined models are randomly shuffled and anonymized, and each sample is evaluated by the following three dimensions:
\begin{itemize}
    \item \textbf{Grammaticality.} It checks whether a question is well-formed. Annotators are asked to rate a sample as 3 for ``no grammatical errors'', 2 for ``not grammatically correct but able to infer actual meaning'', and 1 for ``unacceptable".
    \item \textbf{Answerability.} As elaborated by \citet{nema-khapra-2018-towards}, this dimension checks whether a question is answerable according to the presence and correctness of important information such as named entities, content (relation) words, and question types. Annotators are asked to rate a sample as 3 for ``all important information is present'', 2 for ``some important information is missing'', and 1 for ``all important information is missing''.
    \item \textbf{Relevance.} Following the human criteria used in QG-STEC Task B~\cite{rus-etal-2010-first}, this dimension checks whether a question is consistent with the context and can be answered by the given answer span correctly. Annotators are asked to rate a sample as 2 for ``Completely relevant to the context and given answer'' and 1 for ``totally irrelevant''.
\end{itemize}
In addition to the detailed annotation instruction, the annotators were also informed of the clear scoring examples as summarized in Table~\ref{tab:scoring-examples}.
As shown in Figure~\ref{fig:software}, we develop a web application to collect the evaluation results automatically.
The software will provide candidate questions to the human annotators, guide them to perform annotation, and post their ratings back to our server.
After that, we can analyze the final human judgments based on the results on our server.

\begin{table}[!t]
    \centering
     \resizebox{1.0\columnwidth}{!}{
        \begin{tabular}{c|ccc|ccc|ccc}
            \toprule
            \multicolumn{1}{c}{\multirow{2}{*}{\textbf{Metrics}}}& \centerthreeg{\textbf{Grammaticality}}& \centerthreeb{\textbf{Answerability}}& \centerthreer{\textbf{Relevance}} \\
            \multicolumn{1}{c}{}& \centeroneg{\textbf{$r$}}& \centeroneg{\textbf{$\rho$}}& \centeroneg{\textbf{$\tau$}}& \centeroneb{\textbf{$r$}}& \centeroneb{\textbf{$\rho$}}& \centeroneb{\textbf{$\tau$}}& \centeroner{\textbf{$r$}}& \centeroner{\textbf{$\rho$}}& \centeroner{\textbf{$\tau$}} \\
            \midrule
            \midrule
                BLEU-4& 0.117& 0.108& 0.108& 0.165& 0.146& 0.112& 0.104& 0.078& 0.076 \\
                ROUGE-L& 0.150& 0.142& 0.141& 0.194& 0.175& 0.136& 0.131& 0.107& 0.099 \\
                METEOR& 0.164& 0.155& 0.154& 0.208& 0.188& 0.147& 0.143& 0.120& 0.109 \\
                \hline
                Q-BLEU& 0.280& 0.272& 0.270& 0.314& 0.293& 0.233& 0.243& 0.225& 0.193 \\
                BERTScore& 0.316& 0.309& 0.305& 0.347& 0.327& 0.258& 0.272& 0.257& 0.219 \\
                MoverScore& 0.334& 0.326& 0.323& 0.363& 0.344& 0.273& 0.289& 0.273& 0.230 \\
                BLEURT& 0.354& 0.346& 0.341& 0.380& 0.359& 0.287& 0.304& 0.289& 0.244 \\
                COMET& 0.409& 0.401& 0.395& 0.429& 0.408& 0.328& 0.349& 0.338& 0.282 \\
                \hline
                Q-BLEU$_{free}$& 0.344& 0.332& 0.331& 0.369& 0.349& 0.278& 0.295& 0.279& 0.235 \\
                BERTScore$_{free}$& 0.379& 0.371& 0.367& 0.402& 0.384& 0.306& 0.324& 0.313& 0.260 \\
                CTC& 0.410& 0.404& 0.401& 0.430& 0.412& 0.329& 0.353& 0.340& 0.284 \\
                BARTScore& 0.417& 0.408& 0.406& 0.440& 0.420& 0.334& 0.359& 0.346& 0.289 \\
                \hline
                \underline{QRelScore}& \underline{0.460}& \underline{0.453}& \underline{0.449}& \underline{0.478}& \underline{0.460}& \underline{0.366}& \underline{0.397}& \underline{0.386}& \underline{0.319} \\ 
                \textbf{Ref-QRelScore}& \textbf{0.480}& \textbf{0.473}& \textbf{0.468}& \textbf{0.495}& \textbf{0.474}& \textbf{0.379}& \textbf{0.412}& \textbf{0.404}& \textbf{0.332} \\
            \bottomrule
        \end{tabular}
     }
    \caption{Segment-level correlation in terms of Pearson’s $r$, Spearman’s $\rho$, and
    Kendall’s $\tau$ with human judgments on the HotpotQA dataset.
    The best and second-best results are \textbf{bold} and \underline{underlined}, respectively.}
    \label{tab:segment-correlation-hotpotqa}
\end{table}

\begin{figure}[!t]
	\centering

	\begin{subfigure}{0.49\columnwidth}
    \includegraphics[width=\textwidth]{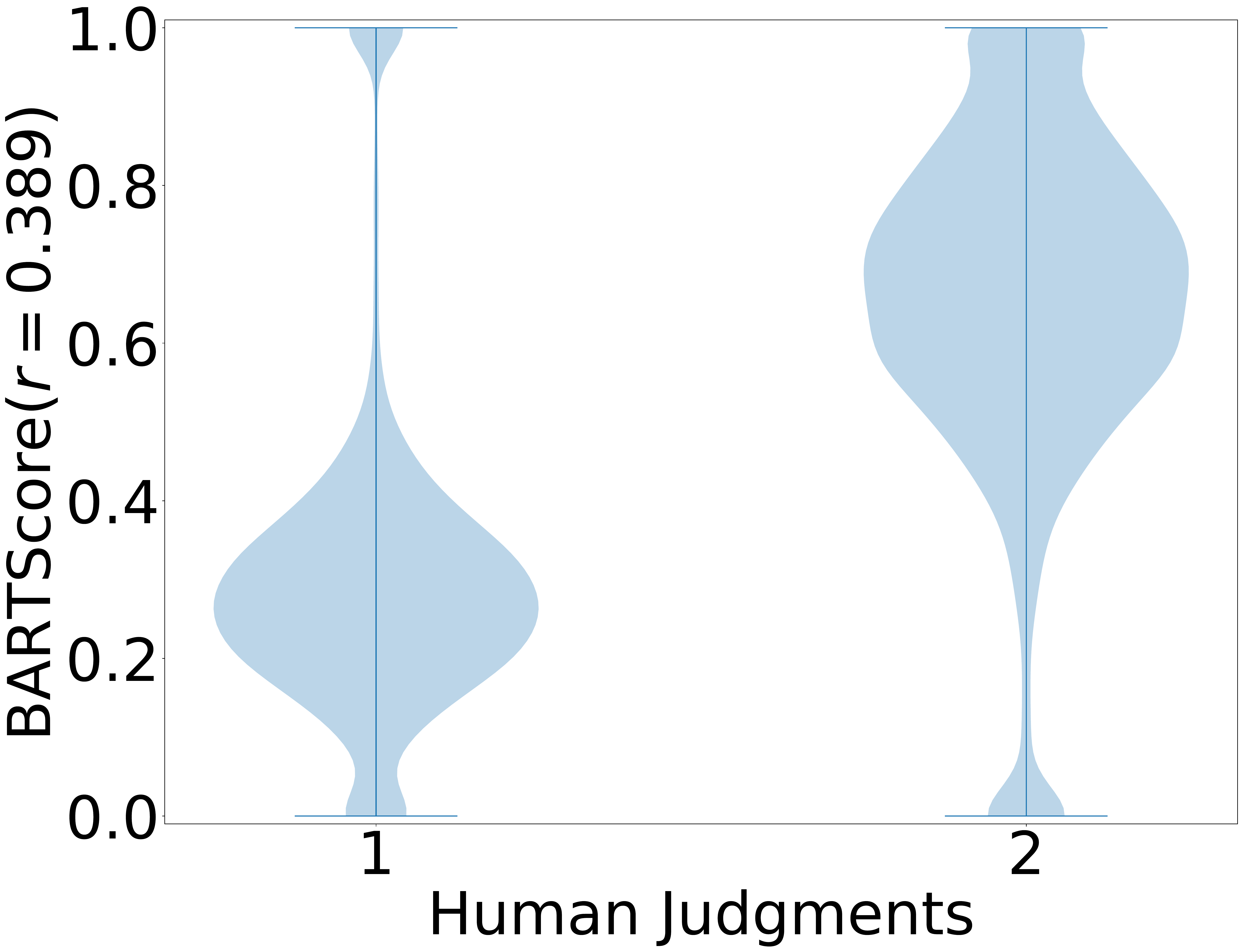}
	\end{subfigure}
	\hfill
	\begin{subfigure}{0.49\columnwidth}
    \includegraphics[width=\textwidth]{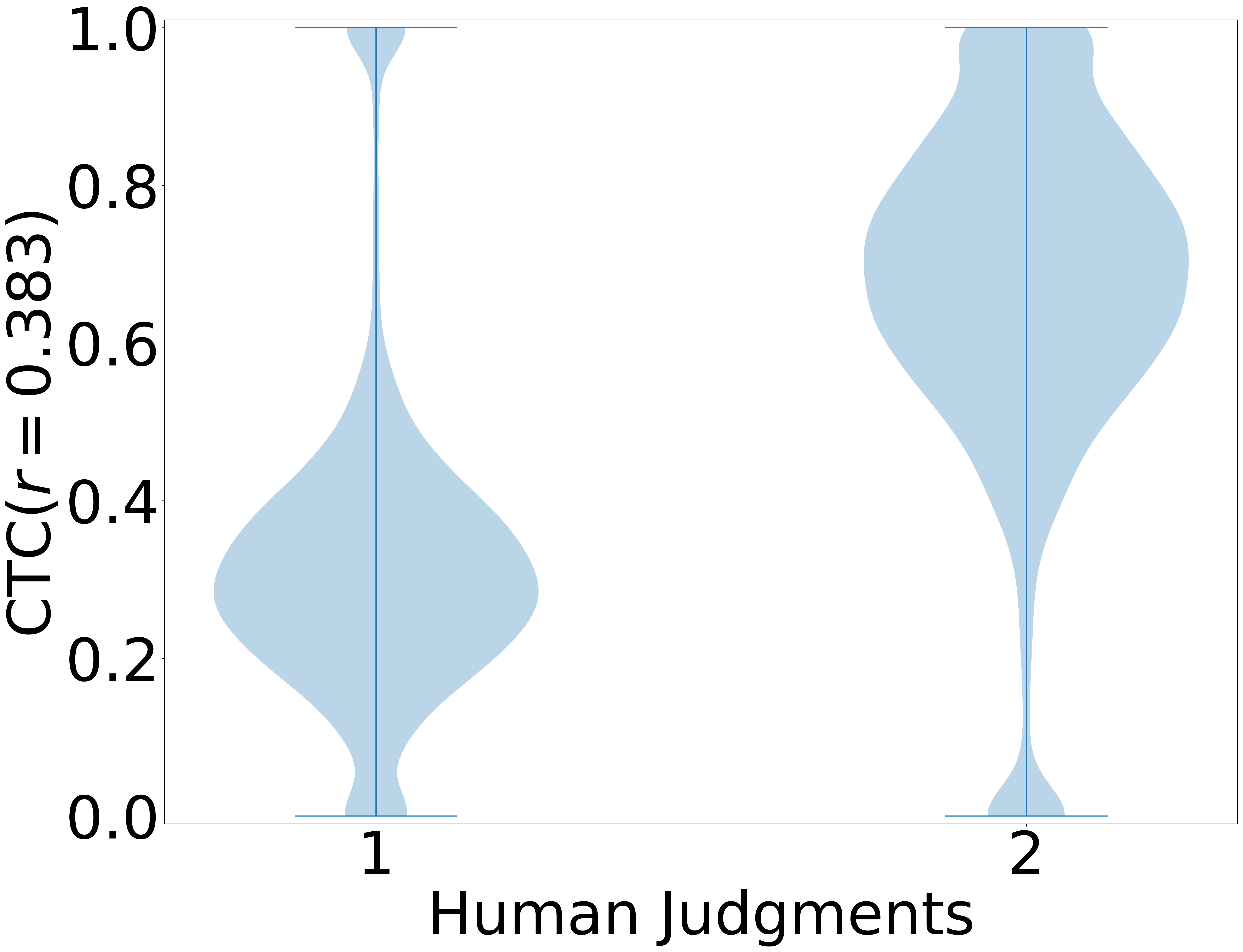}
	\end{subfigure}
    \\
	\begin{subfigure}{0.49\columnwidth}
    \includegraphics[width=\textwidth]{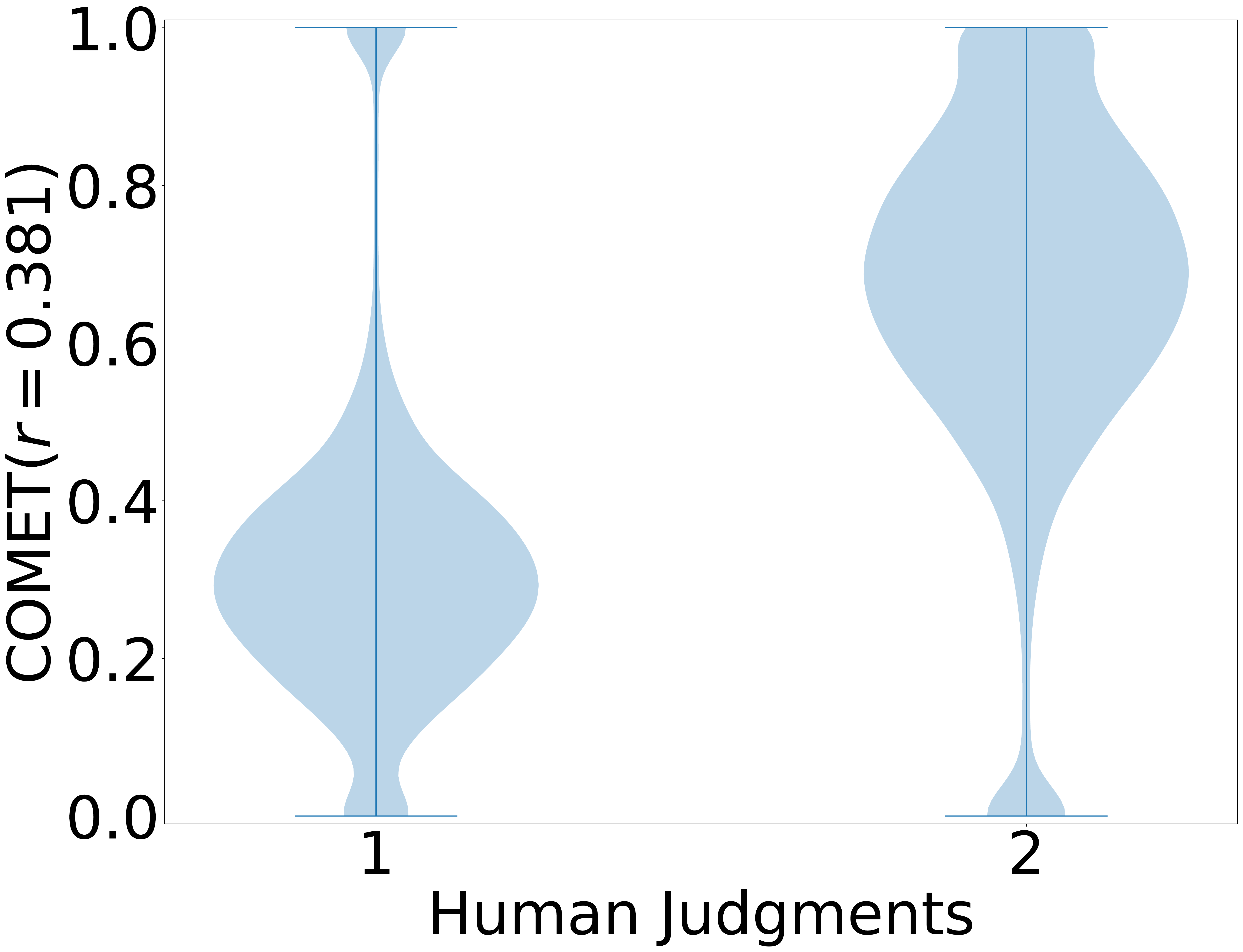}
	\end{subfigure}
	\hfill
	\begin{subfigure}{0.49\columnwidth}
    \includegraphics[width=\textwidth]{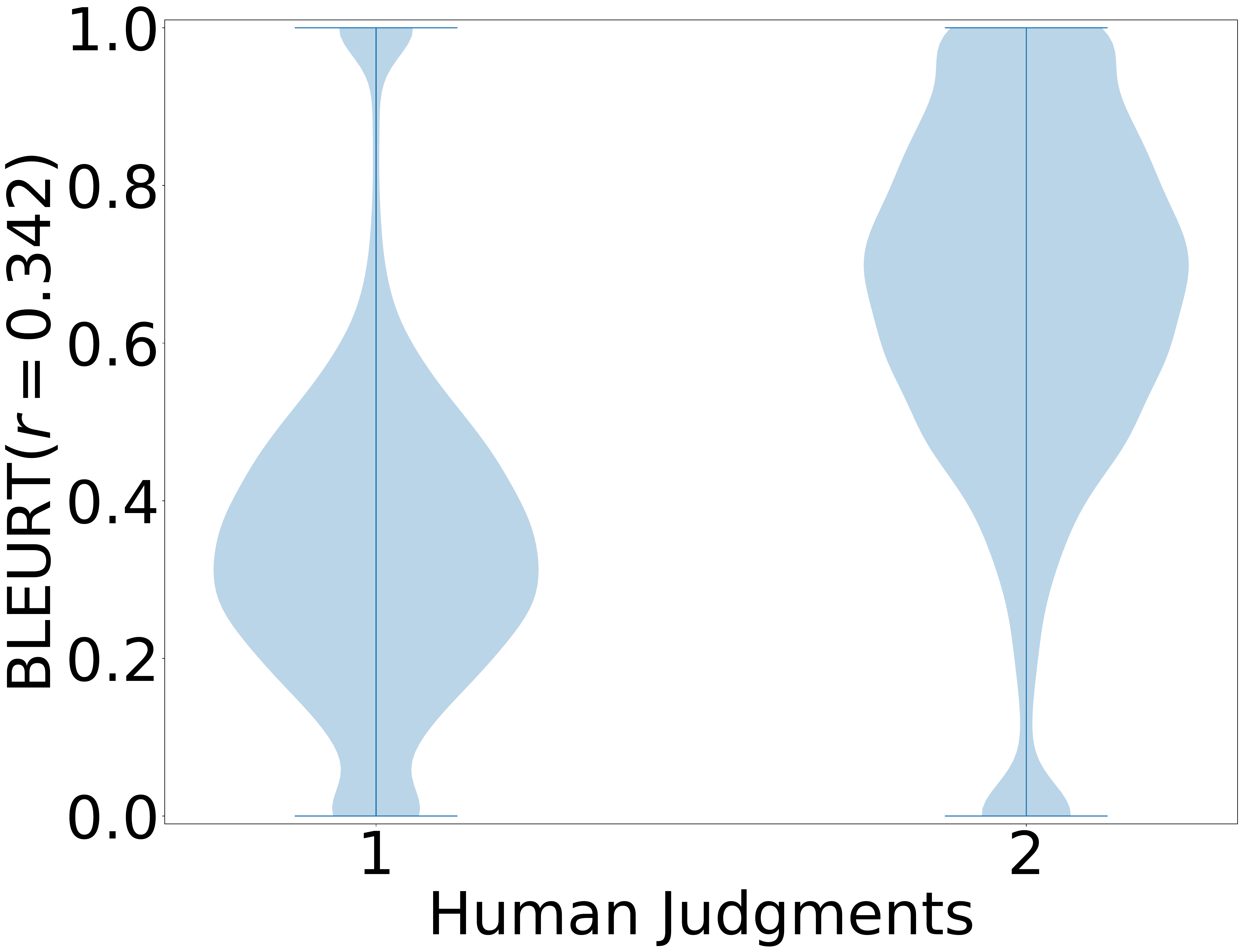}
	\end{subfigure}
    \\
	\begin{subfigure}{0.49\columnwidth}
    \includegraphics[width=\textwidth]{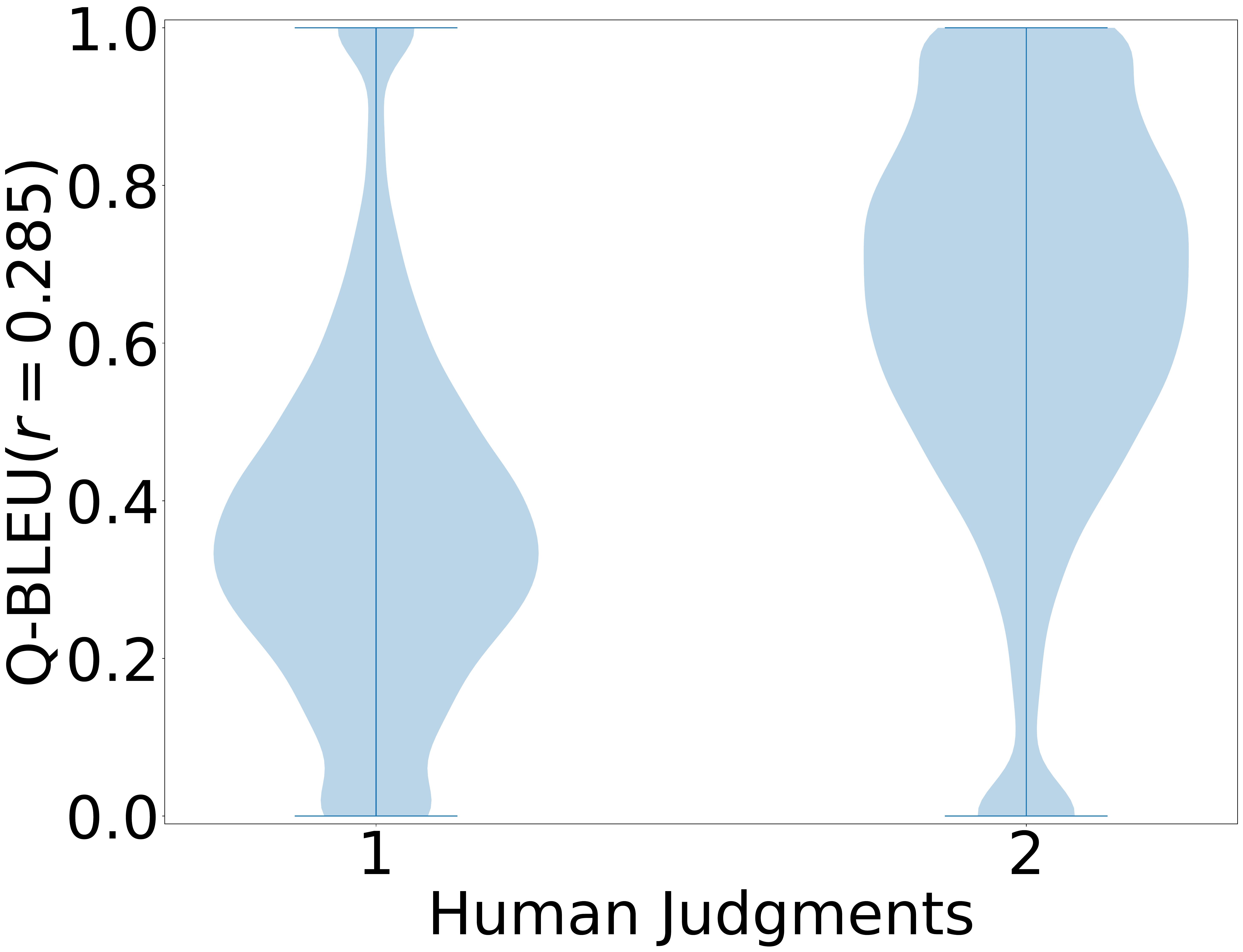}
	\end{subfigure}
	\hfill
	\begin{subfigure}{0.49\columnwidth}
    \includegraphics[width=\textwidth]{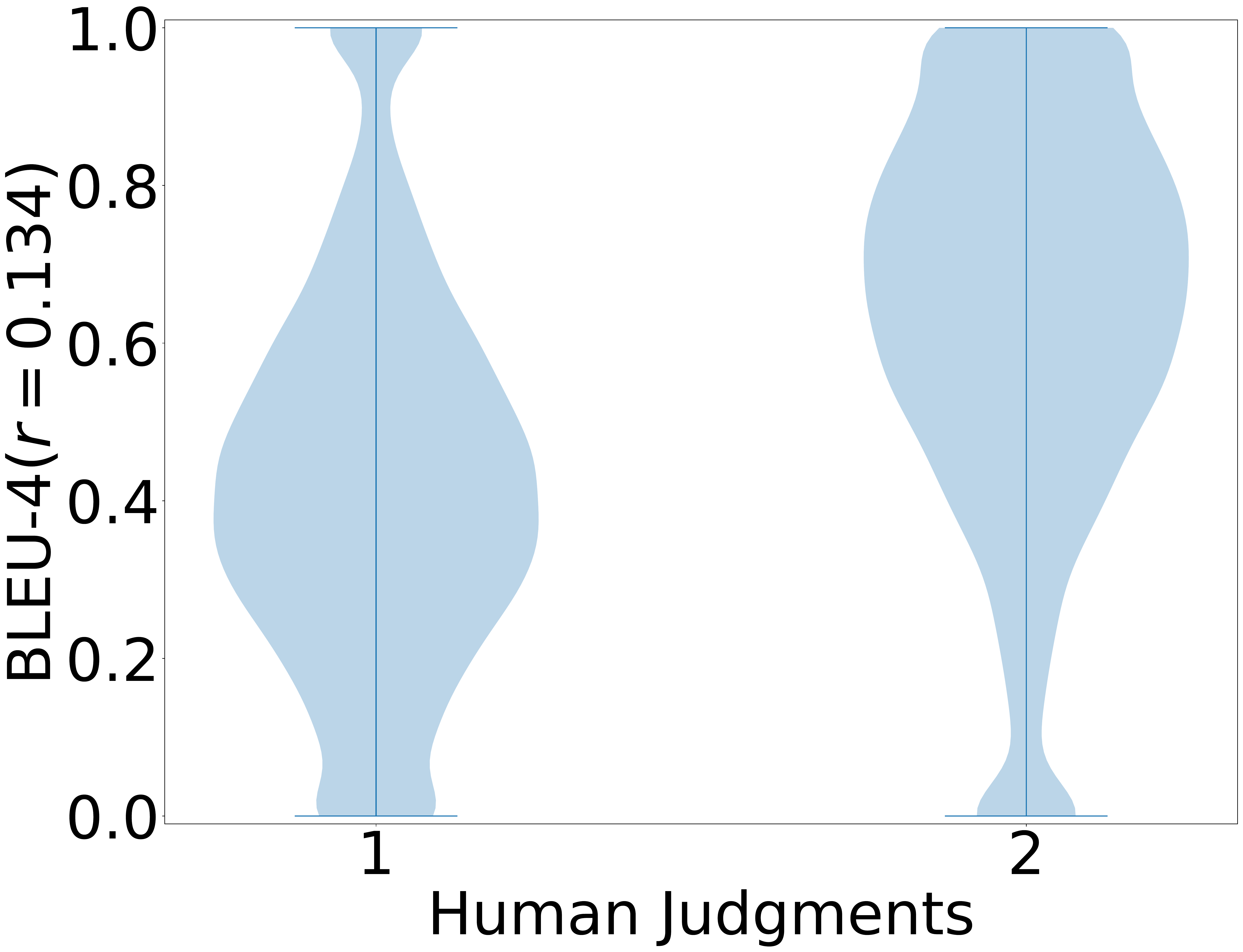}
	\end{subfigure}

    \caption{Score distributions of BARTScore, CTC, COMET, BLEURT, Q-BLEU and BLEU-4 under different relevance ratings (\ie 1 for ``relevant'' and 2 for ``irrelevant'') of human judgments.}
     \label{fig:more-score-distribution}
\end{figure}

\section{More Experimental Results}
\label{sec:more-experiments}

Table~\ref{tab:segment-correlation-hotpotqa} presents
the segment-level correlation to human judgments on the HotpotQA dataset.
We observe that our QRelScore consistently outperforms all the baselines, which indicates the effectiveness of incorporating language models into the relevance evaluation of QG.

Figure~\ref{fig:more-score-distribution} qualitatively illustrates the score distributions of COMET, BLEURT, Q-BLEU, and BLEU-4 under different relevance ratings of human judgments.
These metrics poorly correlate with human judgments because they either assign relatively low scores to the candidates of high quality or score the candidates of a certain level of quality with high variance.

\begin{table*}[!t]
    \centering
     \resizebox{1.0\textwidth}{!}{
        \begin{tabular}{l|l|lll}
            \toprule
            \multicolumn{2}{c}{\textbf{Instruction}}& \textbf{Context}& \textbf{Candidate question} \\
            \hline
            \multirow{6}{*}{\rotatebox[origin=c]{90}{\textbf{Grammaticality}}}& \multirow{2}{5cm}{3 = No grammatical errors}& \multirow{2}{8cm}{$[\ldots]$ Denver linebacker Von Miller was named Super Bowl MVP, recording \textcolor{dgreen}{five} solo tackles $[\ldots]$}& \multirow{2}{8cm}{How many solo tackles did Von Miller make at Super Bowl 50?} \\
            & & & \\
            \cline{2-4}
            & \multirow{2}{5cm}{2 = Not grammatically correct but able to infer actual meaning}& \multirow{2}{8cm}{$[\ldots]$ Miami's Sun Life Stadium and the San Francisco Bay Area's \textcolor{dgreen}{Levi's Stadium} $[\ldots]$}& \multirow{2}{8cm}{What site is \textcolor{dred}{locate} in the San Francisco Bay Area?} \\
            & & & \\
            \cline{2-4}
            & \multirow{2}{5cm}{1 = Unacceptable grammaticality}& \multirow{2}{8cm}{$[\ldots]$ Kubiak replacing \textcolor{dgreen}{Elway} at the end of the Broncos' defeats in Super Bowls XXI $[\ldots]$}& \multirow{2}{8cm}{\textcolor{dred}{Why} was replaced \textcolor{dred}{of} Kubiak in Super Bowl XXIV?} \\
            & & & \\
            \hline
            \multirow{6}{*}{\rotatebox[origin=c]{90}{\textbf{Answerability}}}& \multirow{2}{5cm}{3 = All important information is present}& \multirow{2}{8cm}{$[\ldots]$ \textcolor{dgreen}{six-time} Grammy winner and Academy Award nominee Lady Gaga $[\ldots]$}& \multirow{2}{8cm}{How many Grammys has Lady Gaga won?} \\
            & & & \\
            \cline{2-4}
            & \multirow{2}{5cm}{2 = Some important information is missing}& \multirow{2}{8cm}{$[\ldots]$ and one of the largest in East-Central Europe, employing \textcolor{dgreen}{2,000} professors $[\ldots]$}& \multirow{2}{8cm}{How many \textcolor{dred}{\st{professors}} does the Warsaw University of Technology employ?} \\
            & & & \\
            \cline{2-4}
            & \multirow{2}{5cm}{1 = All important information is missing}& \multirow{2}{8cm}{$[\ldots]$ liberated by \textcolor{dgreen}{Napoleon's} army in 1806, Warsaw was made the capital $[\ldots]$}& \multirow{2}{8cm}{Whose \textcolor{dred}{\st{army liberated}} Warsaw in \textcolor{dblue}{\st{1806}}?} \\
            & & & \\
            \hline
            \multirow{4}{*}{\rotatebox[origin=c]{90}{\textbf{Relevance}}}& \multirow{2}{5cm}{2 = Completely relevant to the context and given answer}& \multirow{2}{8cm}{$[\ldots]$ \textcolor{dgreen}{the Vistula River} is the specific axis of Warsaw, which divides the city into two parts $[\ldots]$}& \multirow{2}{8cm}{What is the axis of Warsaw which divides it into two parts?} \\
            & & & \\
            \cline{2-4}
            \cline{2-4}
            & \multirow{2}{5cm}{1 = Totally irrelevant}& \multirow{2}{8cm}{$[\ldots]$ transmitting mechanical energy with minimal loss over \textcolor{dgreen}{any terrestrial distance} $[\ldots]$}& \multirow{2}{8cm}{\textcolor{dred}{Who received a bid in 1935?}} \\
            & & & \\
            \bottomrule
        \end{tabular}
     }
    \caption{Human annotation instructions along with the scoring examples for the grammaticality, answerability, and relevance dimension.
    The given answers and problematic words in corresponding candidate questions are marked in \textcolor{dgreen}{green} and \textcolor{dred}{red}, respectively.
    }
    \label{tab:scoring-examples}
\end{table*}

\section{Adversarial Examples}
\label{sec:adversarial-details}

\begin{table*}[!t]
    \centering
     \resizebox{1.0\textwidth}{!}{
        \begin{tabular}{l|ll}
            \toprule
            \multicolumn{1}{l}{\textbf{Transformation}}& \textbf{Original question}& \textbf{Transformed question} \\
            \hline
            \multirow{2}{*}{Paraphrasing}& \multirow{2}{8cm}{\textcolor{dgreen}{On what date} did the NFL announce that Coldplay would \textcolor{dgreen}{headline the half-time show}?}& \multirow{2}{8cm}{\textcolor{dred}{When} did the NFL announce that Coldplay would \textcolor{dred}{mark the title of the half-time program}?} \\
            & & \\
            \hline
            \multirow{2}{*}{Entity swap}& \multirow{2}{8cm}{Into what language did \textcolor{dgreen}{Marlee Matlin} translate the national anthem?}& \multirow{2}{8cm}{Into what language did \textcolor{dred}{Lady Gaga} translate the national anthem?} \\
            & & \\
            \hline
            \multirow{2}{*}{Pronoun swap}& \multirow{2}{8cm}{In 2005, what did Doctor Who think the condition of \textcolor{dgreen}{his} home planet was?}& \multirow{2}{8cm}{In 2005, what did Doctor Who think the condition of \textcolor{dred}{your} home planet was?} \\
            & & \\
            \hline
            \multirow{2}{*}{Sentence negation}& \multirow{2}{8cm}{What \textcolor{dgreen}{controls} wages in a purely capitalist mode of production?}& \multirow{2}{8cm}{What \textcolor{dred}{doesn't control} wages in a purely capitalist mode of production?} \\
            & & \\
            \bottomrule
        \end{tabular}
     }
    \caption{Examples of text transformations used to generate adversarial samples.
    \textcolor{dgreen}{Green} and \textcolor{dred}{red} text highlight the changes made by the transformation.
    Among these transformations, paraphrasing is a semantically invariant transformation, while sentence negation, entity swap, and pronoun swap are semantically variant transformations.}
    \label{tab:adversarial-examples}
\end{table*}

As shown in Table~\ref{tab:adversarial-examples}, on the one hand, positive samples are constructed by \textbf{paraphrasing transformation}, which is implemented by back-translation with the multi-lingual MarianMTModel~\cite{junczys-dowmunt-etal-2018-marian}.
The original sentence was translated to an intermediate language and translated back to English, yielding a semantically-equivalent sentence with minor syntactic and lexical changes.
French, German, Chinese, Spanish, and Russian were used as intermediate languages. These languages were chosen based on the performance of current NMT systems with the expectation that well-performing languages could ensure better translation quality.
On the other hand, negative samples are generated by the following perturbations:
\begin{itemize}
    \item \textbf{Entity and pronoun swapping.} For entity extraction, a named entity recognition (NER) system is applied to both the reference question and the context to extract all mentioned entities. It divides them into four groups comprising named entities, covering persons, location/institution/organization names, and number entities.
    After that, the random entity sampled from the entity set is swapped within its corresponding group.
    In this work, we use the spaCy NER tagger~\cite{honnibal2017natural}.
    For pronouns, all gender-specific pronouns were first extracted from the reference question.
    Next, a randomly chosen pronoun was swapped with a different one from the same pronoun group to ensure syntactic correctness.
    \item \textbf{Sentence negation.} In the first step, the reference question is scanned in search of auxiliary verbs and modal verbs.
    Then, we randomly choose a verb and add \emph{not} after it or use WordNet~\cite{miller1995wordnet} wrapped in the NLTK~\cite{bird2009natural} package to find its antonym to negate the sentence.
\end{itemize}

\section{Redundancy Analysis}
\label{sec:redundancy-analysis}
Although our QRelScore achieves a better correlation with human judgments than other metrics, it is unclear if individual metrics capture distinct or redundant dimensions of human judgment.
For example, while QRel$_{LRM}$ and BERTScore both produce relatively high correlation, are they redundant or complementary?
This redundancy arises from the difference in the gold-standard input of our QRelScore and other metrics. That is, we use the context as the input while others use the reference, and the content of the reference is usually contained within the corresponding context.
Following \citet{hessel2021clipscore}, we seek a minimal set of metrics that explains the most variance in human judgment and fits it approximately.
To be precise, we undertake a forward selection algorithm~\cite{thompson1995stepwise} on the metrics set consisting of the baselines, our QRelScore, QRel$_{LRM}$ and QRel$_{GRG}$.
This algorithm performs an iterative greedy selection by picking the most informative additional metric from the metrics set and adding it into the target set, which is initially empty.
In this work, we use the implementation of sklearn package~\cite{pedregosa2011scikit} and repeat the forward selection algorithm ten times in 5-fold cross-validation to prevent randomness.

\begin{figure}[!t]
	\centering

	\begin{subfigure}{0.49\columnwidth}
    \includegraphics[width=\textwidth]{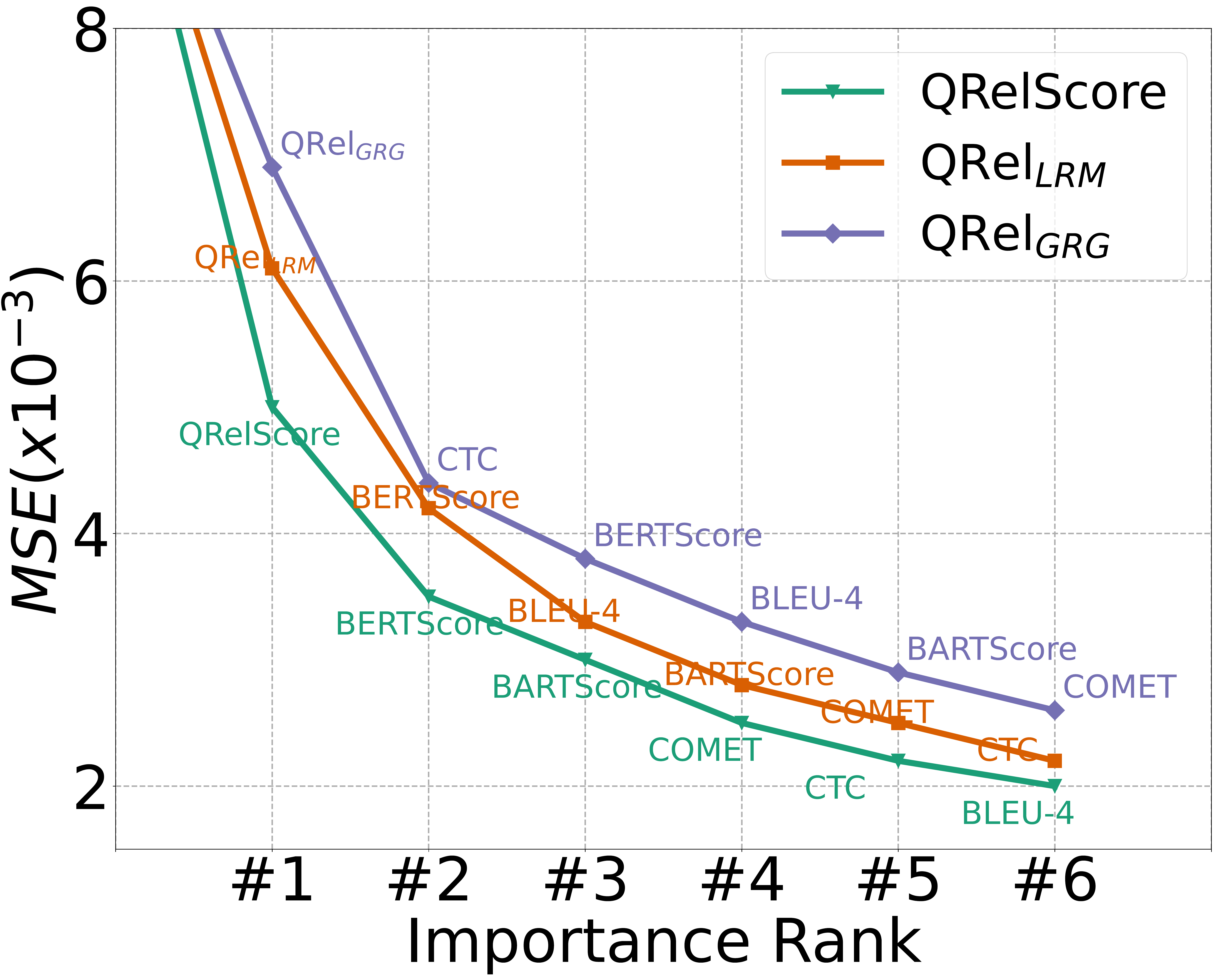}
	\end{subfigure}
	\hfill
	\begin{subfigure}{0.49\columnwidth}
    \includegraphics[width=\textwidth]{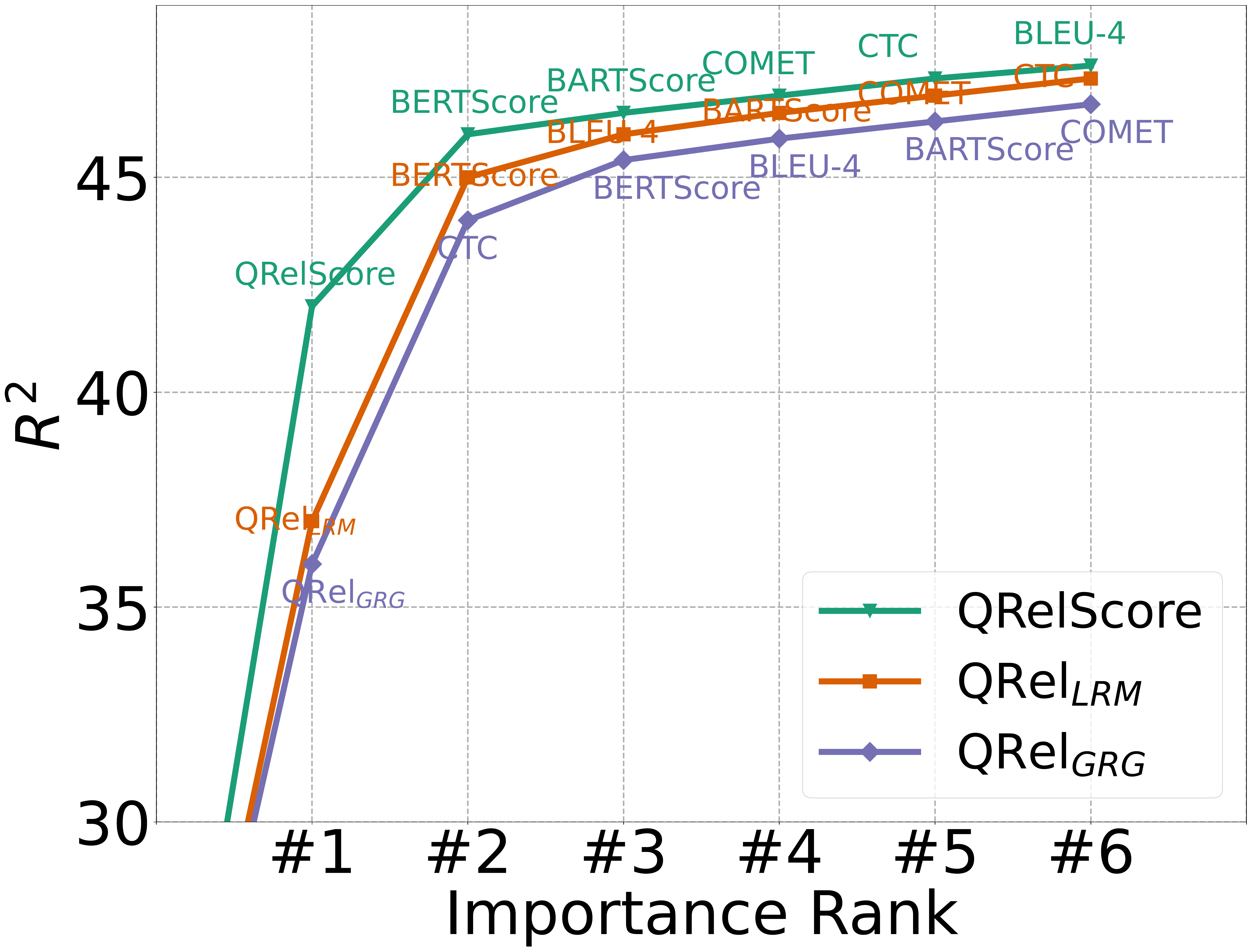}
	\end{subfigure}

    \caption{$MES$ and $R^2$ for the forward-selection regression of metrics on the SQuADv1 dataset.
    Its horizontal axis represents which metric is most commonly chosen at each selection iteration, and a metric that is chosen earlier means more informativeness than the remaining metrics.
    Only the top-6 metrics are illustrated in this diagram.
    }
     \label{fig:redundancy}
\end{figure}

Figure~\ref{fig:redundancy} shows the information gain obtained by different metrics in terms of both mean squared error ($MSE$) and determination coefficient ($R^2$).
On the on hand, we can see that our QRelScore, QRel$_{LRM}$ and QRel$_{GRG}$ tend to be chosen early by the forward selection and make significant improvements to $MSE$ and $R^2$.
This result shows that our reference-free metrics contribute substantial information gain to fitting the human judgments.
On the other hand, reference-based metrics such as BERTScore, BLEU-4, and BLEURT are chosen closely after our reference-free metrics, demonstrating that reference-free evaluation plays a complementary and not redundant role in measuring the overall relevance of QG.

\end{document}